\documentclass[journal]{IEEEtran}
\usepackage{graphicx}
\usepackage{amsmath, amssymb}
\usepackage{soul, xcolor}
\usepackage{caption, subcaption}
\usepackage{hyperref}
\usepackage{cite}
\usepackage{url}
\usepackage{footnote}
\makesavenoteenv{tabular}
\usepackage{multirow}
\usepackage{array}

\begin{document}
\title{3D Fourier-based Global Feature Extraction for Hyperspectral Image Classification}
\author{Muhammad Ahmad
\thanks{M. Ahmad is with SDAIA-KFUPM, Joint Research Center for Artificial Intelligence (JRCAI), King Fahd University of Petroleum and Minerals, Dhahran, 31261, Saudi Arabia. (e-mail: mahmad00@gmail.com).}
}
\markboth{Journal of \LaTeX\ }%
{M.Ahmad \MakeLowercase{\textit{et al.}}}
\maketitle
\begin{abstract}
Hyperspectral image classification (HSIC) has been significantly advanced by deep learning methods that exploit rich spatial-spectral correlations. However, existing approaches still face fundamental limitations: transformer-based models suffer from poor scalability due to the quadratic complexity of self-attention, while recent Fourier transform-based methods typically rely on 2D spatial FFTs and largely ignore critical inter-band spectral dependencies inherent to hyperspectral data. To address these challenges, we propose Hybrid GFNet (HGFNet), a novel architecture that integrates localized 3D convolutional feature extraction with frequency-domain global filtering via GFNet-style blocks for efficient and robust spatial-spectral representation learning. HGFNet introduces three complementary frequency transforms tailored to hyperspectral imagery: Spectral Fourier Transform (a 1D FFT along the spectral axis), Spatial Fourier Transform (a 2D FFT over spatial dimensions), and Spatial-Spatial Fourier Transform (a 3D FFT jointly over spectral and spatial dimensions), enabling comprehensive and high-dimensional frequency modeling. The 3D convolutional layers capture fine-grained local spatial-spectral structures, while the Fourier-based global filtering modules efficiently model long-range dependencies and suppress noise. To further mitigate the severe class imbalance commonly observed in HSIC, HGFNet incorporates an Adaptive Focal Loss (AFL) that dynamically adjusts class-wise focusing and weighting, improving discrimination for underrepresented classes. Additionally, an adaptive fully connected classification head enhances flexibility and generalization across datasets with varying class distributions. Extensive experiments on benchmark hyperspectral datasets demonstrate that HGFNet consistently outperforms state-of-the-art methods in terms of accuracy, effectively overcoming the limitations of prior approaches. The source code will be made publicly available on GitHub.
\end{abstract}
\section{Introduction}

Hyperspectral Image (HSI) Classification (HSIC) is a challenging problem in remote sensing and Earth observation, aiming to assign a semantic land-cover/material label to each pixel by exploiting hundreds of contiguous spectral bands \cite{10472541, 11392780}. Unlike RGB imagery, hyperspectral measurements provide dense reflectance signatures that enable fine-grained discrimination among materials with subtle spectral differences \cite{TULU2026111273, 11105087}, which is essential for precision agriculture \cite{KARUKAYIL2026111282, 11119702}, environmental monitoring \cite{SERRANTI2026129361, 11037730}, mineral exploration \cite{LKHAOUA2026109997, Ahmad18072025}, and land-use/land-cover mapping \cite{10838328}. However, HSIC remains challenging due to the intrinsic properties of hyperspectral data: extremely high dimensionality, strong inter-band correlation (spectral redundancy), and complex spatial heterogeneity \cite{10955699, 10731855}. The ``curse of dimensionality'' manifests in two coupled ways: (i) the effective sample complexity increases rapidly with the feature dimension, and (ii) redundant and noisy bands can dominate learning when labeled samples are limited, which increases the risk of overfitting and unstable decision boundaries \cite{AHMAD2025130428}. In practical scenarios, the difficulty is further amplified by scarce and imbalanced annotations, since ground truth acquisition requires expensive field surveys and expert interpretation \cite{11222092, 10685113}, while classes often exhibit high inter-class spectral similarity (e.g., vegetation species) and large intra-class variability induced by illumination \cite{9903062, Wang26}, atmospheric effects, phenological stages, and mixed pixels \cite{10628006, 9767615}.

Deep learning has substantially advanced HSIC by learning hierarchical spatial-spectral features directly from data \cite{11226902}, with 3D convolutional neural networks (3D CNNs) and transformers emerging as two dominant paradigms \cite{10746459}. 3D CNNs are effective because their kernels jointly convolve across spatial neighborhoods and spectral bands, thereby encoding local spectral gradients, local texture, and neighborhood context in a unified manner \cite{SARPONG2024122202}. This local inductive bias is particularly valuable for hyperspectral imagery, where small spatial structures and local mixing patterns can disambiguate classes with similar spectra \cite{11090003}. Nevertheless, 3D CNNs can be computationally expensive and memory intensive, and their locality limits the receptive field unless deeper stacks or larger kernels are used, which often increases the number of parameters and exacerbates the dependence on larger labeled datasets for stable generalization \cite{10509762}. 

Transformers, in contrast, are attractive due to their ability to capture long-range dependencies through self-attention, enabling global context aggregation that improves spatial consistency and class boundary delineation \cite{10493162, 10399798, 10604879}. Recent hybrid designs attempt to reduce transformer overhead while retaining global modeling capacity, such as MSC-3DLT, which combines multi-scale convolutional extraction with lightweight transformer modules \cite{10850760}, and SSFBF, which couples CNN and transformer components via dynamic routing attention to enhance high-resolution spatial-spectral semantics \cite{10855493}. Despite these advances, transformer-based HSIC pipelines can still be constrained by the quadratic attention cost and by the difficulty of scaling to high-dimensional spectral inputs without careful tokenization and dimensionality reduction.

\begin{figure*}[!hbt]
    \centering
    \includegraphics[width=0.99\linewidth]{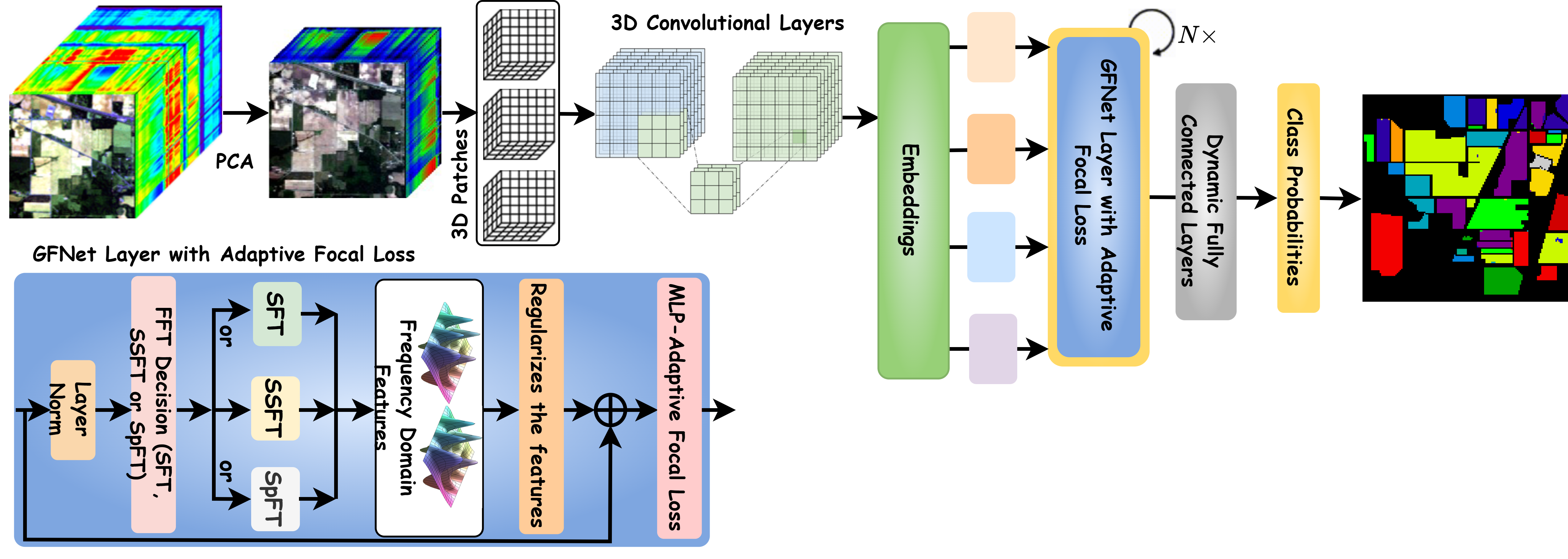}
    \caption{HSI cubes are first divided into 3D patches to preserve local and contextual dependencies. Three consecutive 3D Convolutional layers extract hierarchical features, while the GFNet layer enhances representations by transforming spatial-spectral data into the frequency domain. After normalization, a 3D FFT decomposes features into frequency components, where learnable global filters emphasize relevant information and suppress noise. The transformed features undergo regularization and pass through a Multilayer perceptron (MLP) block with AFL to handle class imbalances. A residual connection ensures stable gradient flow and feature preservation before dynamic fully connected layers generate the final class probabilities.}
    \label{Fig1}
\end{figure*}

Beyond these mainstream approaches, several works incorporate specialized mechanisms to balance local detail and global context. For example, SClusterFormer introduces spatial-spectral cluster attention and deformable convolutions to adaptively model fine-grained structures, though its robustness can be affected by dataset-dependent spectral resolution, sensor noise, and the stability of clustering under limited annotations \cite{10820058}. MCTGCL integrates CNNs, transformers, and graph contrastive learning for planetary HSIC, offering strong accuracy with reduced complexity, but challenges such as domain shift, limited labels, and generalization across acquisition conditions remain significant \cite{10843260}. These limitations motivate exploring alternative representations and operators that can provide efficient global interactions with strong inductive priors suited to hyperspectral data.

Frequency-domain learning has recently emerged as a promising direction to complement and replace attention-based global modeling. In particular, Fourier transform-based operators can provide global receptive fields with favorable computational complexity by leveraging the Fast Fourier Transform (FFT), typically $\mathcal{O}(n\log n)$, compared to the $\mathcal{O}(n^2)$ complexity of full self-attention over $n$ tokens \cite{Ahmad03042025}. The Global Frequency Network (GFNet) exploits FFT-based global filtering to enhance robustness to noise and distortions by modulating representations in the frequency domain \cite{10847802, 10091201}. However, when directly applied to hyperspectral imagery, many existing frequency-based designs perform FFT primarily over spatial dimensions and treat spectral channels as static features, which can underutilize the most discriminative aspect of hyperspectral data: inter-band spectral correlation and wavelength-dependent frequency structure. Moreover, hyperspectral datasets commonly exhibit severe class imbalance, where minority classes have very few labeled pixels and are easily dominated during optimization; despite its practical importance, relatively few HSIC architectures incorporate principled loss designs that explicitly adapt training dynamics to such imbalance.

To address these challenges, this work proposes Hybrid GFNet (HGFNet), a novel architecture that couples a lightweight 3D convolutional stem for localized spatial-spectral feature extraction with GFNet-inspired frequency-domain global filtering to jointly learn local and global representations. The design is motivated by a key observation: hyperspectral discrimination requires both (i) local neighborhood evidence to resolve mixed pixels and boundary ambiguity, and (ii) global context to enforce spatial consistency and capture long-range structural dependencies. First, to overcome the limited spectral modeling of prior frequency-based methods, HGFNet introduces a high-dimensional Fourier generalization tailored to hyperspectral inputs. Specifically, we define three complementary transforms: Spectral Fourier Transform (SFT), a 1D FFT along the spectral axis to encode wavelength-wise frequency variations and inter-band dependencies; Spatial Fourier Transform (SpFT), a 2D FFT over spatial dimensions to capture global spatial patterns and long-range correlations; and Spatial--Spectral Fourier Transform (SSFT), a 3D FFT jointly across spatial and spectral dimensions to model coupled frequency structures that arise from the interaction between material spectra and spatial organization. These operators are embedded within GFNet-style blocks and equipped with learnable frequency masks that adaptively modulate frequency components, enabling the network to suppress high-frequency noise while preserving discriminative structures relevant for classification.

Second, HGFNet performs hybrid local-global feature fusion by combining low-level spatial-spectral cues extracted via 3D convolutions with global frequency-aware features produced by Fourier filtering. This coupling provides a stronger inductive bias than using either mechanism alone: the 3D CNN stabilizes learning under limited labels by enforcing locality and spectral continuity, whereas Fourier filtering offers efficient global mixing that improves large-region coherence and reduces patch-level fragmentation. Third, to explicitly address class imbalance, HGFNet incorporates an Adaptive Focal Loss (AFL) that dynamically adjusts class-wise weighting and focusing behavior. Unlike standard focal loss, which uses a global focusing parameter, AFL adapts the focusing strength per class based on class frequency, thereby emphasizing minority and hard samples without excessively down-weighting well-represented classes. This adaptive optimization strategy complements the proposed frequency-aware representation by improving decision boundaries for underrepresented categories, which is critical for reliable HSIC performance in real-world imbalanced settings. In a nutshell, HGFNet is designed to bridge the gap between local spatial-spectral modeling and efficient global dependency capture by introducing hyperspectral-specific high-dimensional Fourier operators, a principled local-global fusion strategy, and an imbalance-aware adaptive loss enabling robust HSIC.

\section{Proposed Methodology}

Let $\mathbf{X} \in \mathcal{R}^{B \times C \times D \times H \times W}$ be the input HSI data, where $B$ is the batch size, $C$ is the number of spectral channels, $D$ is the depth, and $H$ and $W$ are the spatial height and width of the patches. The goal is to learn a function $f: \mathbf{X} \rightarrow \mathbf{Y}$, where $\mathbf{Y} \in \mathcal{R}^{B \times N}$ represents the class probabilities for $N$ land cover classes as shown in Fig. \ref{Fig1}. A 3D convolutional module is applied to capture the spatial-spectral dependencies as: 

\begin{equation}
    \mathbf{X'} = \sigma(\mathbf{W} \circledast \mathbf{X} + b)
\end{equation}
where $\mathbf{W} \in \mathcal{R}^{F \times C \times D_k \times H_k \times W_k}$ represents the 3D convolutional filter, with $F$ denoting the number of output feature maps, $D_k$, $H_k$, and $W_k$ being the kernel sizes in the spectral and spatial dimensions. The convolution operation $\circledast$ extracts hierarchical features across spectral-spatial domains. The term $b \in \mathcal{R}^{F}$ represents the bias, and $\sigma$ is the Gaussian error linear unit (GeLU) activation function $\sigma(x) = x \cdot \Phi(x)$, where $\Phi(x)$ is the standard Gaussian cumulative distribution function. This activation function helps preserve fine-grained spectral-spatial information. The convolution operation is performed twice to refine feature representations while maintaining coherence.

\subsection{GFNet Block}

The GFNet block leverages frequency-domain representations to efficiently capture long-range dependencies. By employing the 3D-FFT, the spatial-spectral features are decomposed into frequency components, enabling effective feature selection and transformation. Prior to the FFT, feature normalization is applied to stabilize numerical computations and enhance convergence as:

\begin{equation}
    X_{\text{norm}} = \frac{\mathbf{X'} - \mu}{\phi}
\end{equation}
where $\mu$ and $\phi$ denote the mean and standard deviation of the input feature tensor, respectively. The FFT operation transforms the spatial-spectral representation into its frequency-domain equivalent as:

\begin{equation}
    \mathbf{X}_{FT} = \mathcal{F}(X_{\text{norm}}, \text{backward})
\end{equation}
where $\mathcal{F}$ represents the FFT function computed with backward normalization. The `backward` norm ensures that the transformation does not introduce unintended scaling effects by preserving the correct magnitude of the transformed signal. Unlike `ortho` scaling the forward and inverse transforms by $1/\sqrt{N}$, the `backward` normalization guarantees: 

\begin{equation}
    \mathcal{F}^{-1}(\mathcal{F}(X_{\text{norm}}, \text{backward}), \text{backward}) = X_{\text{norm}}
\end{equation}

This formulation ensures that the inverse FFT restores the original feature map without additional scaling adjustments, leading to stable feature propagation and improved training performance. To enhance feature extraction, a frequency-domain mask is applied:

\begin{equation}
    \mathbf{X'}_{FT} = \mathbf{X}_{FT} \odot M
\end{equation}
where $M$ is a binary mask that selectively retains informative frequency components while suppressing noise and redundant signals. The Hadamard product ($\odot$) ensures element-wise filtering, preserving dominant spectral features. The frequency-transformed features are passed through a feed-forward network (FFN) to capture complex interactions:

\begin{equation}
    \mathbf{X}_{FFN} = \sigma(\mathbf{W}_1 \mathbf{X'}_{FT} + b_1)
\end{equation}

\begin{equation}
    \mathbf{X}_{\text{out}} = \mathbf{W}_2 \mathbf{X}_{FFN} + b_2
\end{equation}
where $\mathbf{W}_1 \in \mathcal{R}^{d_h \times d}$ and $\mathbf{W}_2 \in \mathcal{R}^{d \times d_h}$ are learnable weight matrices, and $\sigma$ is a non-linearity that enhances feature expressiveness. To prevent information loss and gradient instability, a residual connection is incorporated: 

\begin{equation}
    \mathbf{X}_{\text{final}} = \mathbf{X}_{\text{norm}} + \mathbf{X}_{\text{out}}
\end{equation}

This formulation ensures efficient feature aggregation while mitigating vanishing gradient issues. Finally, the refined representation is passed through fully connected layers: 

\begin{equation}
    \mathbf{Z}_1 = \sigma(\mathbf{W}_3 \mathbf{X}_{\text{final}} + b_3) ~~\&~~ \mathbf{Z}_2 = \sigma(\mathbf{W}_4 \mathbf{Z}_1 + b_4)
\end{equation}

\begin{equation}
    \mathbf{Y} = \text{softmax}(\mathbf{W}_5 \mathbf{Z}_2 + b_5)
\end{equation}
where $\mathbf{W}_3, \mathbf{W}_4, \mathbf{W}_5$ are trainable weight matrices, and $b_3, b_4, b_5$ are bias terms. The final softmax operation converts the output into class probabilities.

\subsection{AFL and Training}

To address the severe class imbalance in hyperspectral datasets, we adapt the AFL proposed in \cite{10483019}, originally designed for natural image classification, to the hyperspectral domain as follows:

\begin{equation}
L = - \sum_{i = 1}^N \alpha_i (1 - p_i)^{\gamma_i} \log(p_i)
\end{equation}
where, $\alpha_i$ is a class-specific weight accounting for class frequency, $p_i$ is the predicted softmax probability for class $i$, and $\gamma_i$ is a dynamic focusing parameter that emphasizes harder samples. Unlike \cite{10483019}, using globally learned parameters, we define both $\alpha_i$ and $\gamma_i$ adaptively per class to reflect the uneven spectral separability and spatial distribution inherent in HSIC tasks. Additionally, our formulation smoothly integrates with frequency-based GFNet blocks, allowing the model to emphasize discriminative spectral features also for underrepresented classes. The training is done by the Adam optimizer as: 

\begin{equation}
    \theta_{t+1} = \theta_t - \eta \frac{\partial L}{\partial \theta}    
\end{equation}
where $\eta$ is the learning rate and $\theta$ represents model parameters.

\section{Experimental Settings and Results}

To demonstrate the proposed HGFNet model's effectiveness, experiments on three HSI datasets, i.e., Indian Pines (IP), WHU Hi HanChuan (HC), and WHU Hi HongHu (HH) have been performed. The IP dataset is widely used for benchmarking hyperspectral models, whereas the HH and HC datasets provide different real-world scenarios. 

The proposed HGFNet is trained using the Adam optimizer with an initial learning rate of $0.001$ and a weight decay of $1\times10^{-6}$ to mitigate overfitting and stabilize parameter updates. The network is optimized for $50$ epochs with a batch size of $64$, which provides a balanced trade-off between convergence stability and computational efficiency. For all experiments, the datasets are randomly partitioned such that $25\%$ of the labeled samples are used for training, $25\%$ for validation to monitor convergence and prevent overfitting, and the remaining $50\%$ are reserved for testing. This split ensures a sufficiently large and unbiased test set, enabling a reliable assessment of generalization performance under limited labeled data, which is a common and challenging setting in HSIC.

The HGFNet consists of several key components designed for effective feature extraction and classification. It begins with three 3D convolutional layers equipped with GELU activation to capture spatial-spectral features. Following this, four GFNet Blocks are employed, integrating FFTs and MLP layers, along with GELU activation, normalization, and dropout, to enhance feature representation. The classification head features a fully connected classifier that is dynamically initialized based on the feature dimensions, incorporating ReLU activations and dropout layers to improve generalization. To address class imbalance, the model utilizes an AFL with class-balanced weighting, ensuring more robust learning across different classes.

\subsection{Ablation Study}

To rigorously analyze the contribution of each architectural component, we conduct an extensive ablation study focusing on frequency-domain modeling and attention/loss design. The study is designed to answer three key questions: (i) how spatial and spectral frequency components individually contribute to hyperspectral representation learning; (ii) whether joint spatial--spectral frequency modeling yields complementary benefits; and (iii) how the proposed adaptive focal loss (AFL) interacts with frequency-aware representations.

\paragraph{Impact of Spatial and Spectral Frequency Modeling.} We first investigate the effect of selectively enabling spatial and spectral Fourier transforms. The SpFT variant retains only spatial FFT, emphasizing periodic structures and texture regularities in the spatial domain, while discarding spectral correlations. Conversely, SFT preserves only spectral FFT, modeling frequency patterns along the wavelength dimension that capture material-specific signatures and subtle spectral variations. The SSFT configuration applies a full 3D FFT, jointly transforming spatial and spectral dimensions to encode long-range dependencies and cross-dimensional frequency interactions.

As reported in Table \ref{Ablation}, SSFT consistently outperforms both SpFT and SFT across all datasets and metrics. This confirms that spatial and spectral frequency cues are complementary rather than redundant. While spatial frequency information captures structural and contextual patterns, spectral frequency components are critical for discriminating materials with similar spatial appearances but distinct spectral responses. The joint transformation enables the network to model complex couplings between spatial layout and spectral variation, which are otherwise inaccessible when operating in a single domain.

\paragraph{Spectral Dominance in Hyperspectral Data.} A consistent trend is that SFT slightly outperforms SpFT in terms of $\kappa$, OA, and AA on most datasets. This observation aligns well with the nature of hyperspectral imagery, where class separability is largely driven by spectral signatures rather than purely spatial patterns. Frequency analysis along the spectral dimension effectively suppresses noise, enhances discriminative wavelength bands, and captures periodic or smooth spectral behaviors associated with specific land-cover classes. Nevertheless, the performance gap between SFT and SSFT highlights that spectral modeling alone is insufficient, and that spatial frequency context remains essential for robust classification.

\paragraph{Comparison with Attention and Other Frequency-Based Baselines.} Compared to the MHSA baseline, all FFT-based variants demonstrate clear improvements, indicating that explicit frequency-domain modeling provides a stronger inductive bias than attention alone. While MHSA captures global dependencies in the feature space, it lacks an explicit mechanism to disentangle low- and high-frequency components, which are crucial for handling illumination variations, spectral redundancy, and spatial texture complexity.

We further compare SSFT with two representative frequency-domain approaches: Wavelets and GFNet. Although Wavelets introduce multi-scale frequency decomposition and GFNet leverages global filtering, both methods underperform SSFT. This suggests that the proposed 3D FFT formulation is more effective at jointly modeling spatial--spectral correlations, whereas Wavelets rely on fixed basis functions and GFNet primarily emphasizes global spatial filtering without explicitly exploiting spectral frequency structure.

\paragraph{Effect of Loss Design: SFL vs. AFL.} Finally, we examine the interaction between frequency-aware representations and loss design. Replacing the standard focal loss (SFL) with the proposed AFL consistently yields further gains across all datasets. AFL dynamically adjusts class-wise emphasis based on training dynamics, which synergizes well with SSFT features that already enhance class separability in the frequency domain. This combination is particularly beneficial for addressing class imbalance and hard-to-classify samples, leading to more stable optimization and improved generalization.

\begin{table}[!hbt]
    \centering
    \caption{Quantitative comparison of different frequency and attention-based configurations, including SFL (SFL + SSFT) and the proposed AFL (AFL + SSFT).}
    \resizebox{\columnwidth}{!}{\begin{tabular}{c|c|ccccccc} \hline 
    \textbf{Data} & \textbf{Metric} & \textbf{MHSA} & \textbf{Wavelets} & \textbf{GFNet} & \textbf{SpFT} & \textbf{SFT} & \textbf{SFL} & \textbf{AFL} \\ \hline 

    \multirow{3}{*}{\textbf{\rotatebox{90}{IP}}} 
    & \textbf{$\kappa$} & 96.21 & 97.06 & 94.22 & 97.26 & 97.66 & 97.92 & \textbf{98.24} \\ 
    & \textbf{OA} & 96.68 & 98.46 & 92.14 & 97.60 & 97.95 & 96.11 & \textbf{98.45} \\  
    & \textbf{AA} & 95.05 & 95.69 & 93.67 & 96.47 & 96.35 & 96.78 & \textbf{97.16} \\ \hline 

    \multirow{3}{*}{\textbf{\rotatebox{90}{HC}}} 
    & \textbf{$\kappa$} & 97.49 & 95.51 & 95.73 & 97.75 & 97.83 & 98.67 & \textbf{99.10} \\
    & \textbf{OA} & 97.85 & 97.38 & 96.48 & 98.07 & 98.15 & 98.87 & \textbf{99.23} \\
    & \textbf{AA} & 95.75 & 97.76 & 94.12 & 96.73 & 96.44 & 97.92 & \textbf{98.71} \\ \hline 

    \multirow{3}{*}{\textbf{\rotatebox{90}{HH}}} 
    & \textbf{$\kappa$} & 97.87 & 96.28 & 95.91 & 98.22 & 98.75 & 98.96 & \textbf{99.13} \\
    & \textbf{OA} & 98.31 & 97.71 & 96.89 & 98.59 & 99.01 & 97.14 & \textbf{99.31} \\
    & \textbf{AA} & 96.53 & 96.64 & 95.02 & 96.76 & 97.97 & 96.12 & \textbf{98.29} \\ \hline 
    \end{tabular}}
    \label{Ablation}
\end{table}

In a nutshell, the ablation results demonstrate that joint SSFT is crucial for maximizing HSIC performance, as it effectively captures complementary spatial structures and discriminative spectral patterns. While spectral frequency information plays a dominant role due to the intrinsic characteristics of hyperspectral data, spatial frequency cues remain essential for providing contextual and structural support. Furthermore, FFT-based modeling consistently outperforms attention-only mechanisms and alternative frequency-domain baselines, indicating the advantage of explicitly disentangling and exploiting frequency components. Finally, the proposed AFL further amplifies these gains when coupled with SSFT by improving class-wise optimization and robustness to hard samples. As a result, AFL + SSFT achieves the best overall performance, reaching $99.13$ $\kappa$ and $99.31$ OA on the HH dataset, thereby validating the effectiveness of the proposed design.

\subsection{Experimental Results and Discussion}

To validate the effectiveness of the proposed HGFNet, we conduct extensive experiments on three benchmark hyperspectral datasets and compare against a diverse set of recent state-of-the-art methods, including SpectralFormer \cite{9627165}, Hybrid ViT (HViT) \cite{khan2024deep}, Wavelet-based spatial-spectral transformer (WaveFormer) \cite{10399798}, wavelet-based spatial-spectral Mamba (WaveMamba) \cite{10767233}, and Spatial-spectral Mamba \cite{10604894}. All methods are trained and evaluated under the same experimental protocol on a single RTX 4060 GPU (8GB VRAM) with 64GB RAM. The goal of this comparative study is to assess not only accuracy (Overall accuracy (OA), Average accuracy (AA), and Kappa coefficient ($\kappa$)) but also the robustness of spatial structures and class boundaries in the predicted maps, thereby highlighting the benefits of combining local 3D CNN inductive biases with global Fourier-domain feature mixing in a unified framework.

\begin{table}[!hbt]
    \centering
    \caption{Class-wise accuracy and overall performance metrics on \textbf{HC Dataset}.} 
    \resizebox{\columnwidth}{!}{\begin{tabular}{c||ccccc||c} \hline 
        \textbf{Class} & SF & HViT & WFormer & Mamba & WMamba & \textbf{HGFNet}  \\ \hline 
        1 & 98.4307 & 99.3383 & 98.4755 & 96.9197 & 96.0255 & 99.7183 \\ 
        2 & 99.2001 & 98.1629 & 99.2528 & 83.8006 & 96.0007 & 99.6396 \\ 
        3 & 94.3807 & 98.6000 & 85.4754 & 78.0283 & 94.8668 & 99.7472 \\ 
        4 & 99.8879 & 99.5517 & 97.9454 & 95.6294 & 99.2528 & 99.7385 \\ 
        5 & 97.8333 & 99.1666 & 97.1666 & 73.0000 & 63.8333 & 99.6666 \\ 
        6 & 92.6301 & 86.4518 & 81.5975 & 35.3045 & 76.8314 & 96.1606 \\ 
        7 & 97.7303 & 94.9186 & 98.4078 & 88.5501 & 95.0203 & 99.9322 \\ 
        8 & 94.8492 & 95.0606 & 96.9073 & 76.1820 & 93.5254 & 98.5426 \\ 
        9 & 96.1765 & 98.2044 & 97.8876 & 57.0553 & 89.5014 & 98.0566 \\ 
        10 & 99.4674 & 98.5545 & 99.4294 & 93.7618 & 95.2073 & 99.9239 \\ 
        11 & 97.7764 & 98.8290 & 97.6818 & 79.5742 & 96.0023 & 96.4163 \\ 
        12 & 84.2934 & 96.7934 & 95.4347 & 58.0978 & 65.4347 & 99.5108 \\ 
        13 & 93.7033 & 90.3905 & 94.5809 & 56.6037 & 72.9486 & 97.6305 \\ 
        14 & 98.2974 & 98.2112 & 98.6206 & 74.2780 & 95.6465 & 99.2564 \\ 
        15 & 87.6760 & 86.4436 & 90.1408 & 62.5000 & 68.4859 & 95.5985 \\ 
        16 & 99.6923 & 99.4111 & 99.8779 & 99.1644 & 98.9177 & 99.9310 \\ \hline \hline 
        \textbf{$\kappa$} & 97.4911 & 97.6654 & 97.3873 & 83.9145 & 93.6020 & \textbf{99.1074} \\ 
        \textbf{OA} & 97.8557 & 98.0048 & 97.7680 & 86.3021 & 94.5280 & \textbf{99.2373} \\ 
        \textbf{AA} & 95.7516 & 96.1305 & 95.5551 & 75.5281 & 87.3438 & \textbf{98.7168} \\ \hline        
    \end{tabular}}
    \label{Tab2}
\end{table}
\begin{figure}[!htb]
    \centering
    \begin{minipage}{0.060\textwidth}
        \centering
        \includegraphics[width=\textwidth]{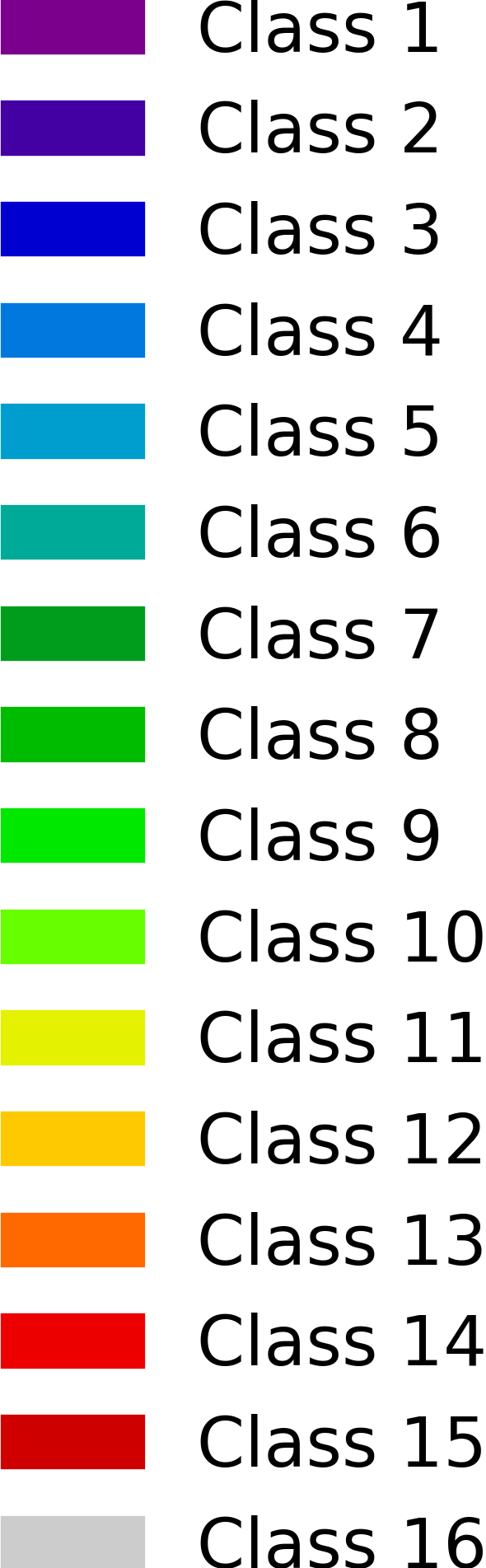}
    \end{minipage}
    \begin{minipage}{0.41\textwidth}
        \centering
        \begin{subfigure}{0.49\textwidth}
            \includegraphics[width=\linewidth]{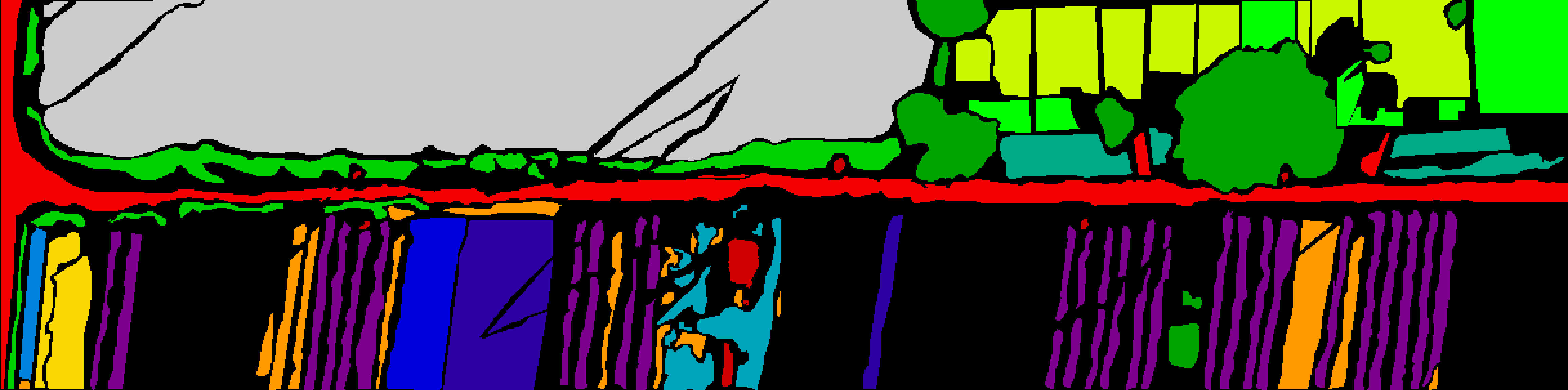}
            \caption{GT}
        \end{subfigure}
        \begin{subfigure}{0.49\textwidth}
            \includegraphics[width=\linewidth]{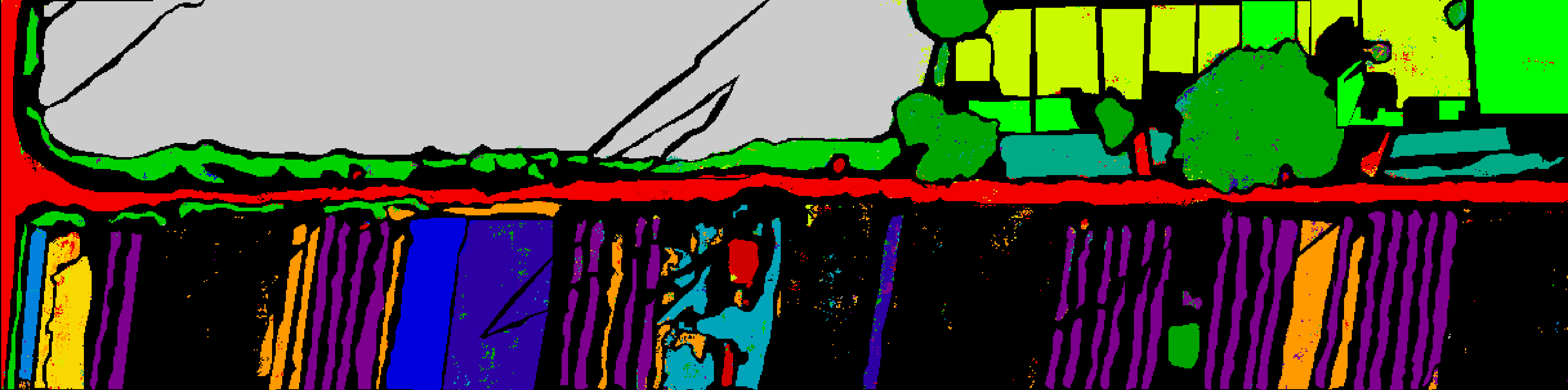}
            \caption{SF}
        \end{subfigure}
        \begin{subfigure}{0.49\textwidth}
            \includegraphics[width=\linewidth]{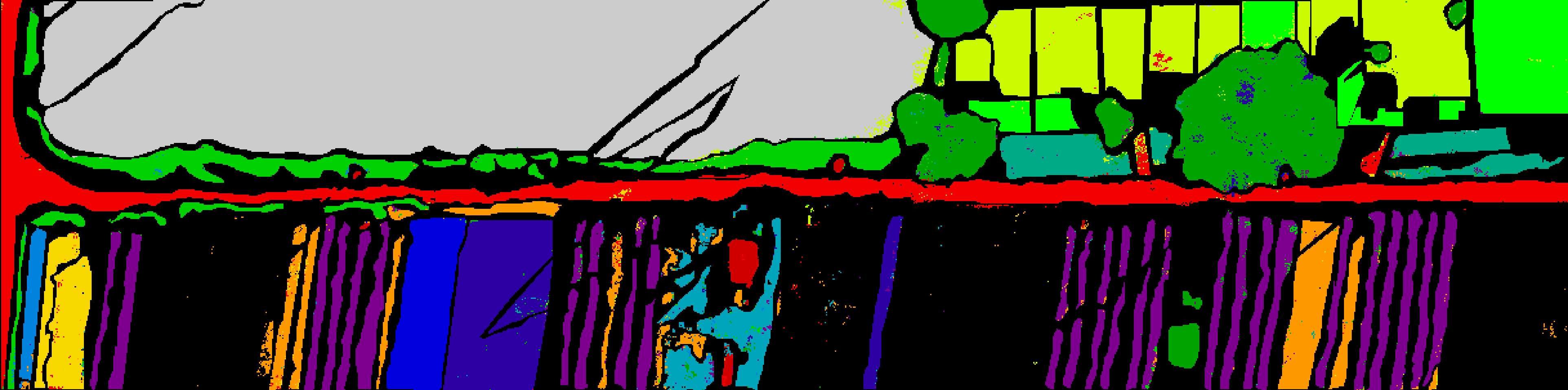}
            \caption{HViT}
        \end{subfigure}
        \begin{subfigure}{0.49\textwidth}
            \includegraphics[width=\linewidth]{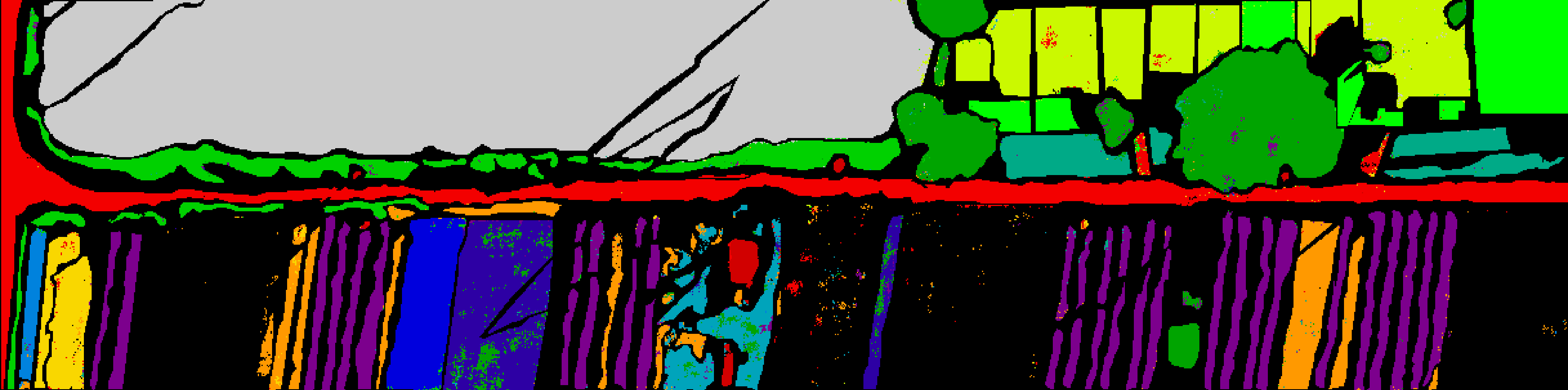}
            \caption{WFormer}
        \end{subfigure}
        \begin{subfigure}{0.49\textwidth}
            \includegraphics[width=\linewidth]{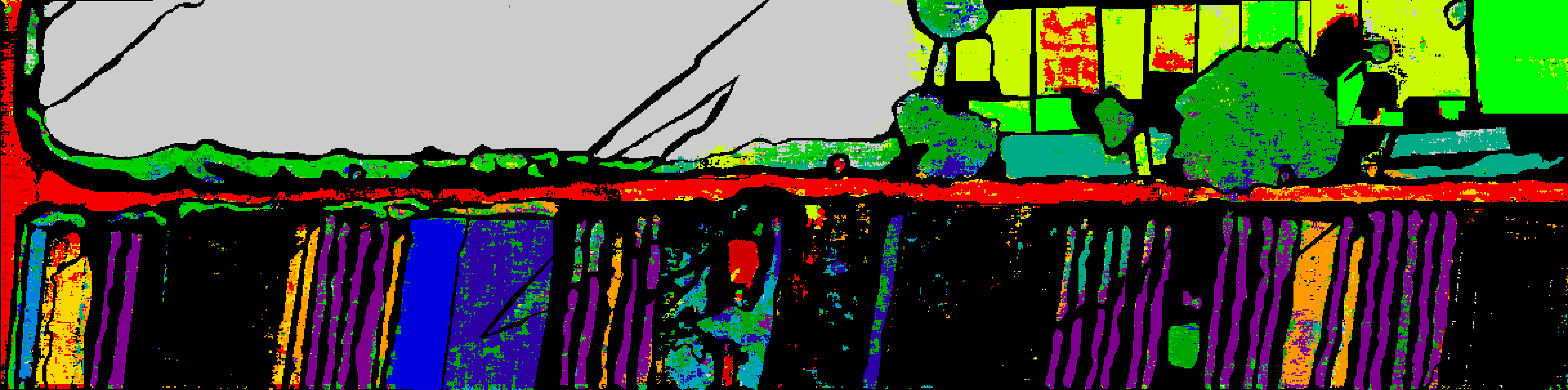}
            \caption{Mamba}
        \end{subfigure}
        \begin{subfigure}{0.49\textwidth}
            \includegraphics[width=\linewidth]{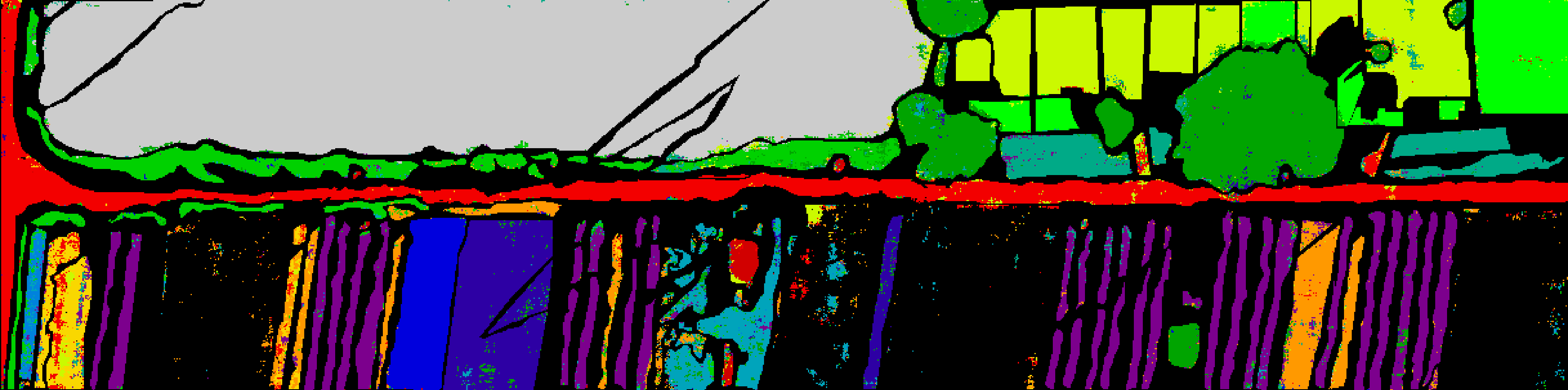}
            \caption{WMamba}
        \end{subfigure}
        \begin{subfigure}{0.49\textwidth}
            \includegraphics[width=\linewidth]{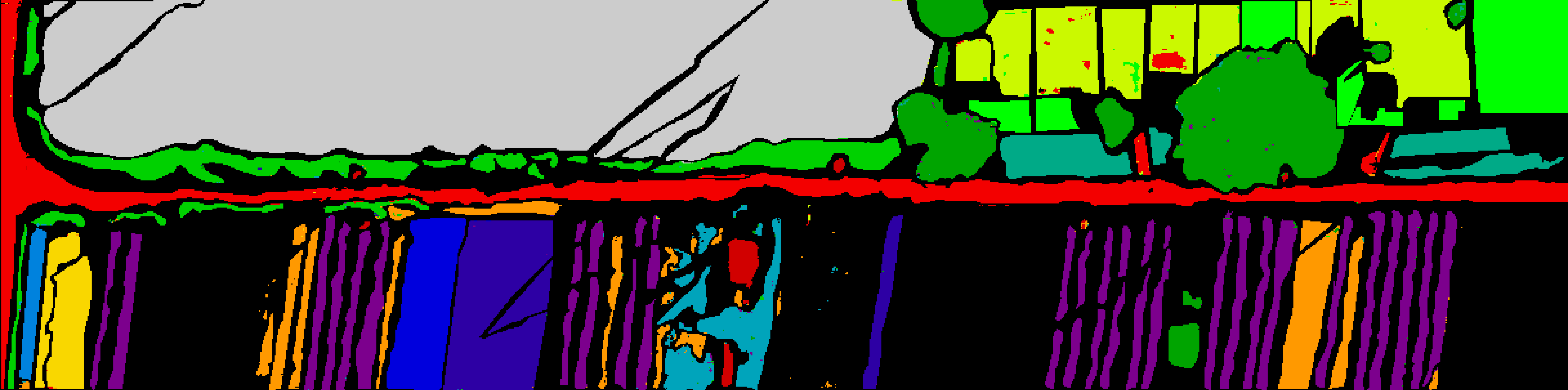}
            \caption{HGFNet}
        \end{subfigure}
    \end{minipage}
    \caption{Visual comparison of classification maps on the HC dataset. (a) Ground truth maps: (b–g) predicted maps from competing models and our proposed HGFNet.}
    \label{HCGT}
\end{figure}

Table \ref{Tab2} summarizes the class-wise accuracies and overall metrics on the HC dataset. HGFNet achieves the best overall performance with an OA of $99.2373\%$, AA of $98.7168\%$, and $\kappa$ of $99.1074\%$, outperforming all competitors across nearly all classes. The performance gain is especially pronounced in classes where strong spectral similarity and intra-class variability typically confuse; in these cases, HGFNet benefits from (i) the local spatial-spectral coupling captured by the 3D CNN front-end, which preserves fine-grained neighborhood context and reduces salt-and-pepper noise, and (ii) the global frequency mixing, which aggregates long-range dependencies and harmonizes globally consistent structures. In contrast, the pure Mamba baseline shows the weakest overall performance (OA $86.3021\%$, $\kappa$ $83.9145$), suggesting that a sequence-modeling bias alone may be insufficient to reliably capture the complex joint spatial--spectral statistics in this dataset, particularly under limited labeled samples and pronounced class imbalance. The qualitative maps in Fig. \ref{HCGT} further corroborate these quantitative trends: HGFNet produces more spatially coherent regions with clearer boundaries and fewer isolated misclassified pixels, indicating that the learned representation is not only discriminative in feature space but also spatially consistent in the image domain.

\begin{table}[!hbt]
    \centering
    \caption{Class-wise accuracy and overall performance metrics on \textbf{IP Dataset}.} 
    \resizebox{\columnwidth}{!}{\begin{tabular}{c||ccccc||c} \hline 
        \textbf{Class} & SF & HViT & WFormer & Mamba & WMamba & \textbf{HGFNet}  \\ \hline 
        1 & 95.6521 & 100 & 86.9565 & 4.3478 & 65.2173 & 67.3913 \\ 
        2 & 95.0980 & 92.5770 & 93.1372 & 42.2969 & 81.5126 & 95.4481 \\ 
        3 & 97.1084 & 96.6265 & 98.3132 & 18.0722 & 77.3493 & 99.6385 \\ 
        4 & 94.0677 & 85.5932 & 91.5254 & 3.3898 & 50.8474 & 100 \\ 
        5 & 97.1074 & 99.5867 & 98.7603 & 33.0578 & 84.7107 & 97.9296 \\ 
        6 & 97.8082 & 98.6301 & 98.0821 & 86.5753 & 87.9452 & 100 \\ 
        7 & 85.7142 & 100 & 100 & 0 & 57.1428 & 100 \\ 
        8 & 100 & 100 & 100 & 93.3054 & 99.5815 & 100 \\ 
        9 & 80 & 90 & 80 & 0 & 60 & 100 \\ 
        10 & 90.7407 & 90.9465 & 90.7407 & 33.5390 & 75.7201 & 99.8971 \\ 
        11 & 98.2899 & 98.7785 & 98.1270 & 81.9218 & 91.2866 & 98.2484 \\ 
        12 & 95.9595 & 96.2962 & 95.9595 & 31.3131 & 69.3602 & 97.1332 \\ 
        13 & 98.0392 & 100 & 100 & 46.0784 & 76.4705 & 100 \\ 
        14 & 98.4202 & 99.2101 & 98.5781 & 92.4170 & 92.8909 & 99.6837 \\ 
        15 & 96.8911 & 97.9274 & 96.3730 & 36.2694 & 73.5751 & 100 \\ 
        16 & 100 & 97.8260 & 97.8260 & 54.3478 & 69.5652 & 100 \\ \hline \hline 

        \textbf{$\kappa$} & 96.2159 & 96.1922 & 96.0382 & 51.1942 & 81.3875 & \textbf{98.2419} \\ 
        \textbf{OA} & 96.6829 & 96.6634 & 96.5268 & 58.3414 & 83.7268 & \textbf{98.4583} \\ 
        \textbf{AA} & 95.0560 & 96.4999 & 95.2737 & 41.0582 & 75.8235 & \textbf{97.1643} \\ \hline 
    \end{tabular}}
    \label{Tab1}
\end{table}
\begin{figure}[!htb]
    \centering
    \begin{minipage}{0.040\textwidth}
        \centering
        \includegraphics[width=\textwidth]{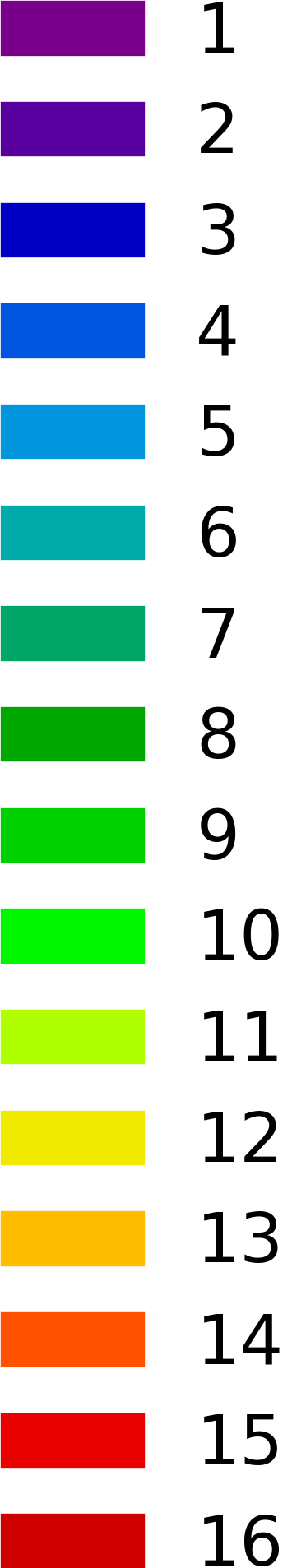}
    \end{minipage}
    \begin{minipage}{0.42\textwidth}
        \centering
        \begin{subfigure}{0.23\textwidth}
            \centering
            \includegraphics[width=0.99\textwidth]{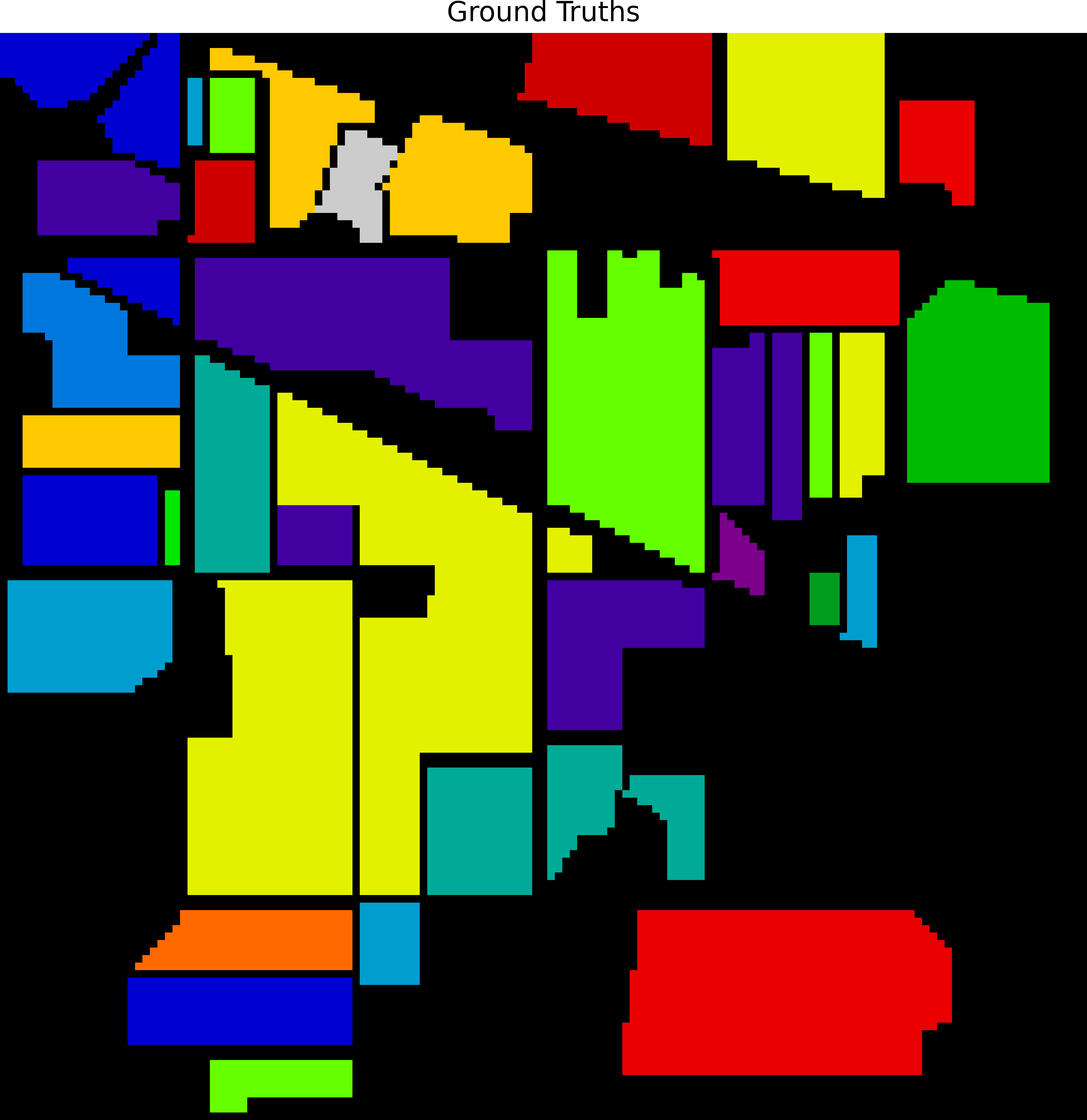}
            \caption*{GT}
        \end{subfigure}
        \begin{subfigure}{0.23\textwidth}
            \centering
            \includegraphics[width=0.99\textwidth]{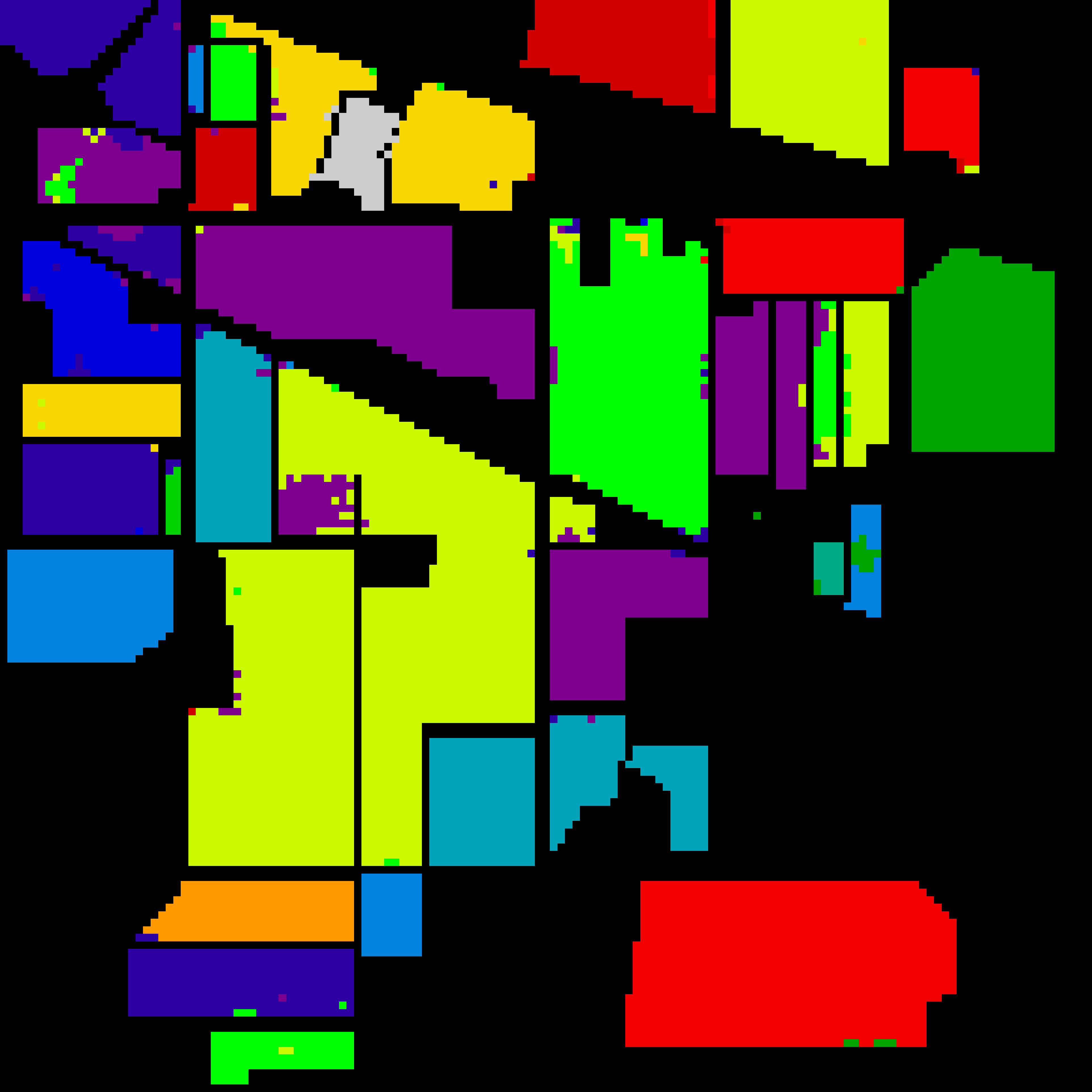}
            \caption*{SF}
        \end{subfigure}
        \begin{subfigure}{0.23\textwidth}
            \centering
            \includegraphics[width=0.99\textwidth]{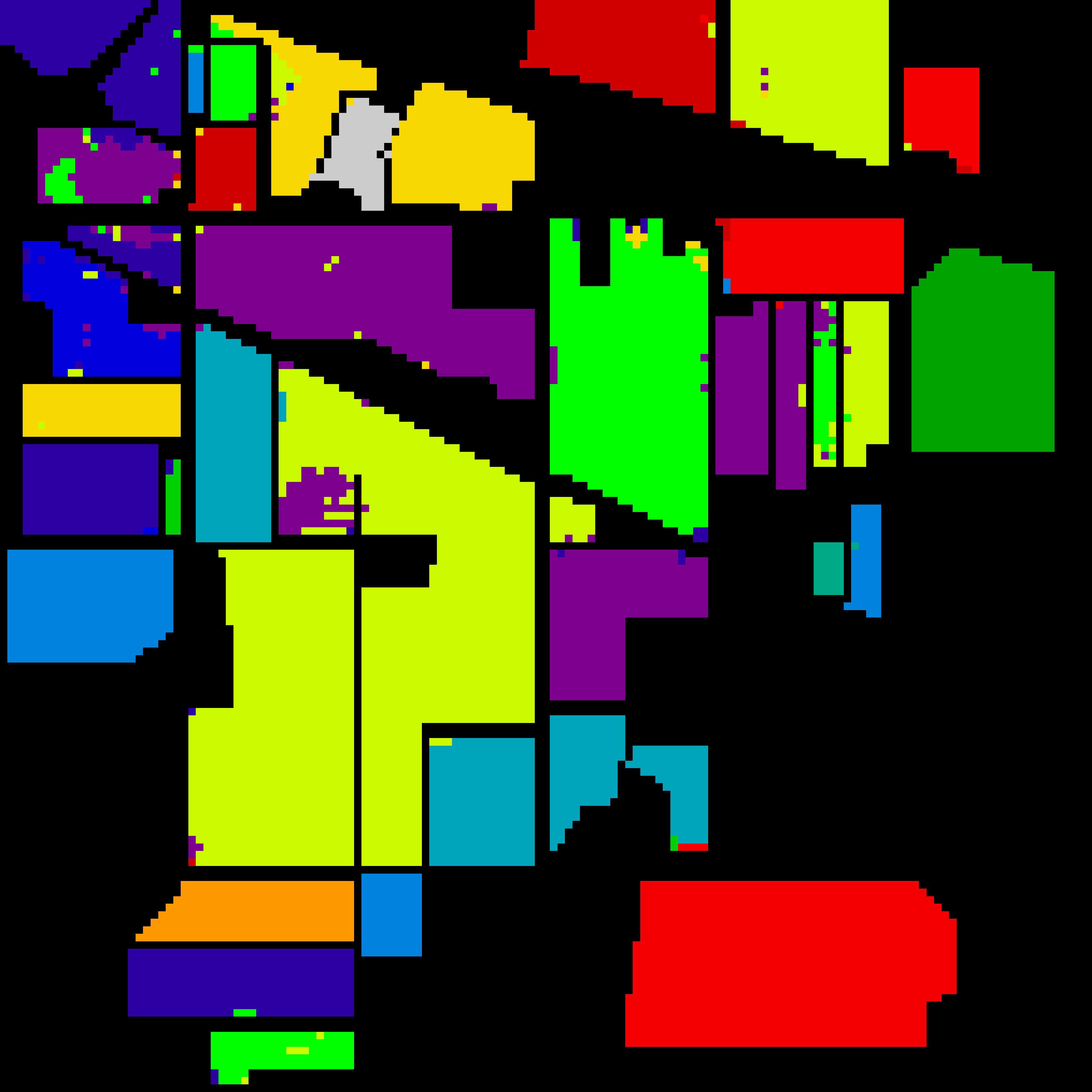}
            \caption*{HViT}
        \end{subfigure}
        \begin{subfigure}{0.23\textwidth}
            \includegraphics[width=0.99\textwidth]{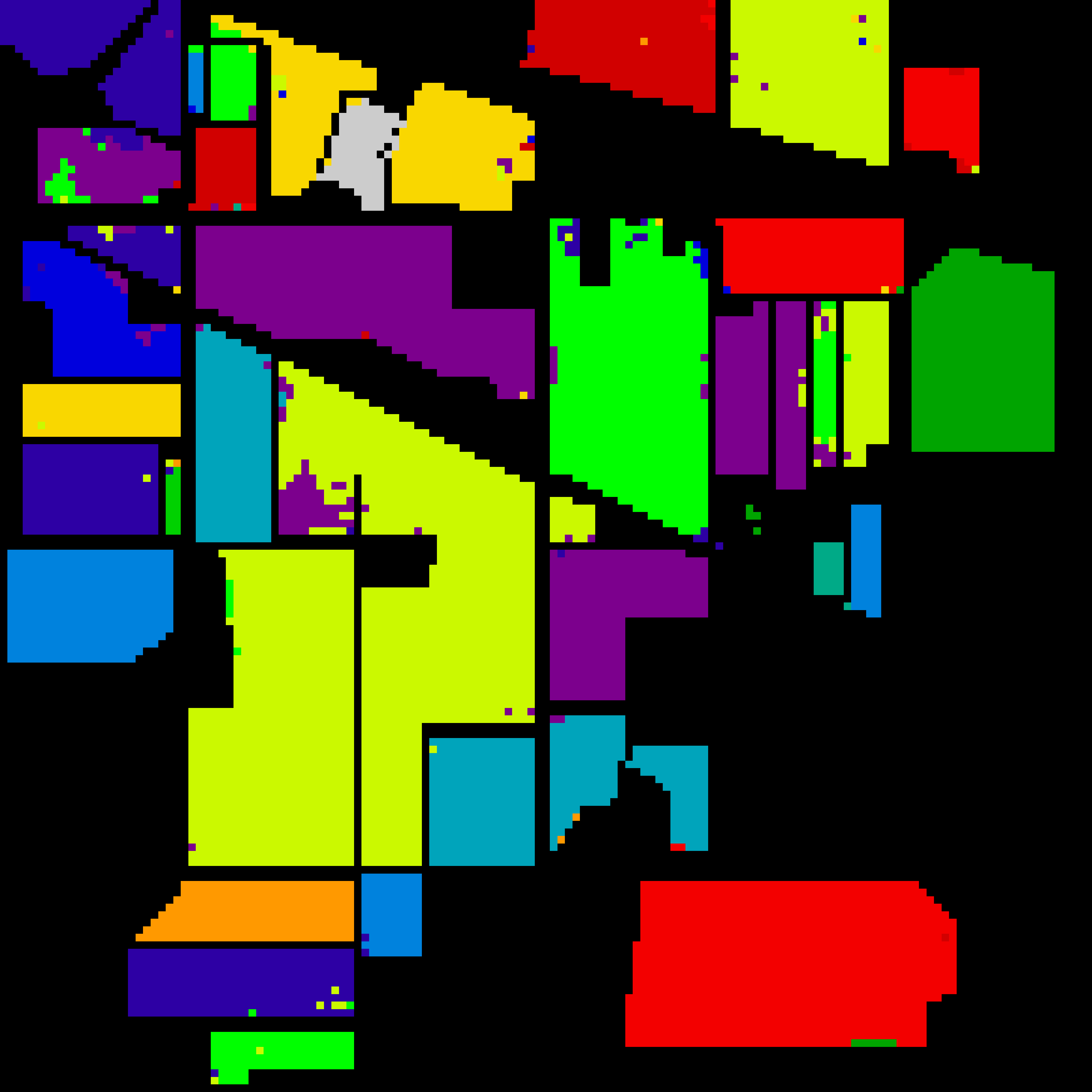}
            \caption*{WFormer}
        \end{subfigure}
        \begin{subfigure}{0.23\textwidth}
            \centering
            \includegraphics[width=0.99\textwidth]{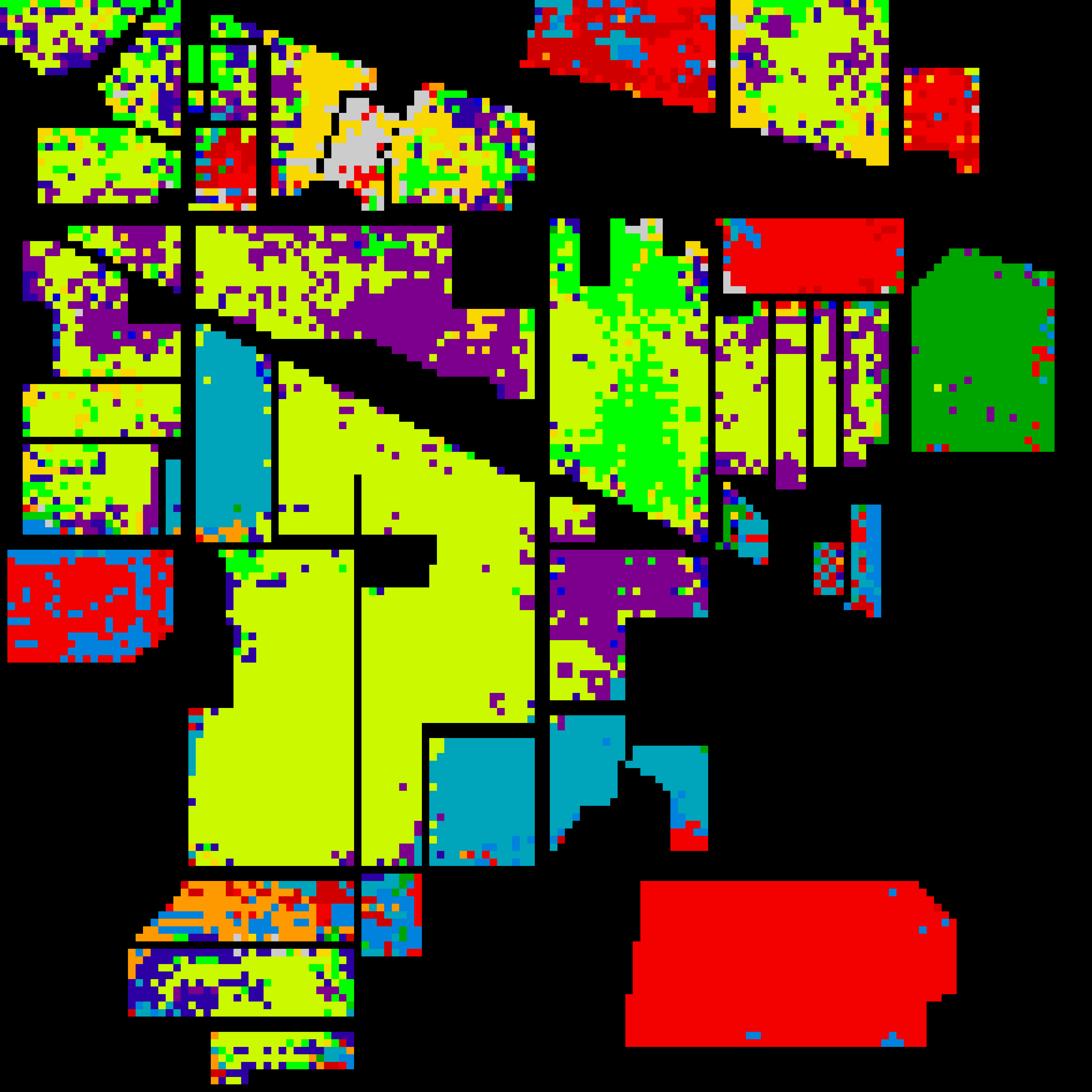}
            \caption*{Mamba}
        \end{subfigure}
        \begin{subfigure}{0.23\textwidth}
            \centering
            \includegraphics[width=0.99\textwidth]{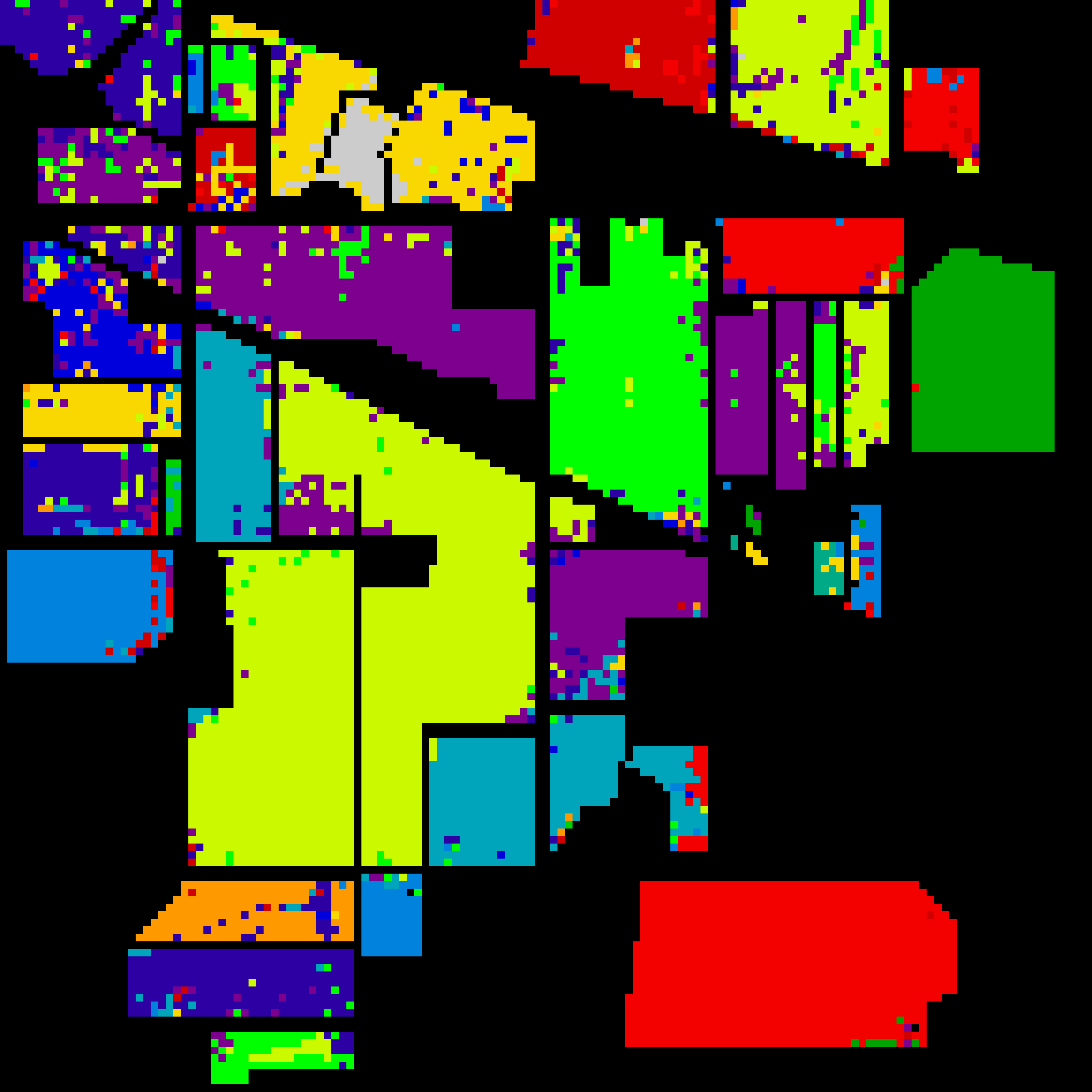}
            \caption*{WMamba}
        \end{subfigure}
        \begin{subfigure}{0.23\textwidth}
            \centering
            \includegraphics[width=0.99\textwidth]{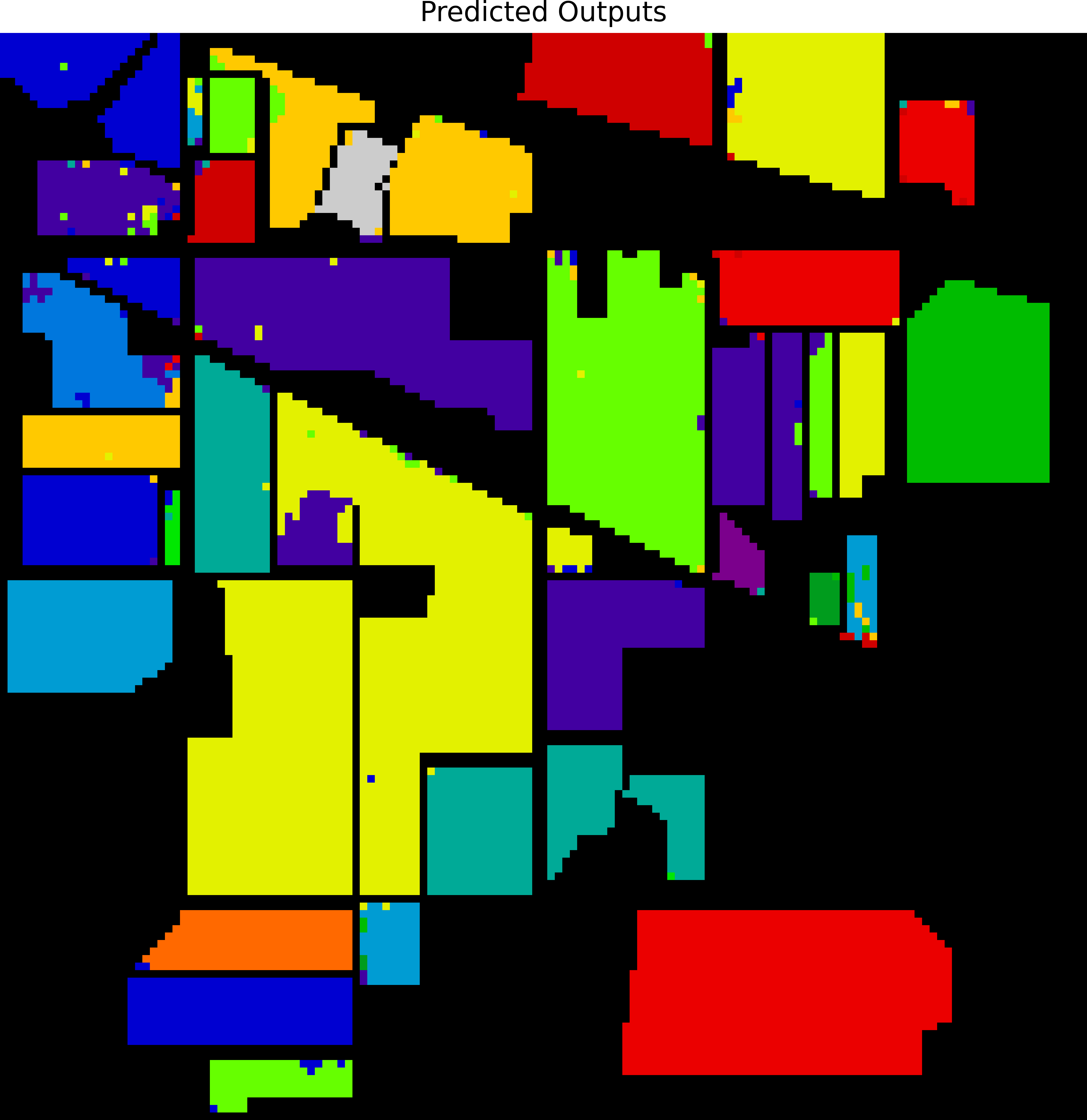}
            \caption*{HGFNet}
        \end{subfigure}
        \end{minipage}
    \caption{Predicted ground truth maps for the Indian Pines dataset.}
    \label{IPGT}
\end{figure}

On the IP dataset (Table \ref{Tab1}), HGFNet again delivers the strongest overall results, reaching an OA of $98.4583\%$, AA of $97.1643\%$, and $\kappa$ of $98.2419\%$. IP is known to contain many spectrally similar vegetation classes and fragmented spatial distributions, which typically amplify confusion between visually adjacent categories and increase boundary errors. The improvements of HGFNet over transformer-based and wavelet-based baselines suggest that its design is particularly effective at simultaneously addressing (i) local ambiguity, by leveraging 3D convolutions to encode local spatial context together with spectral continuity, and (ii) global context dependency, by using Fourier-domain operations to enable efficient global interactions that stabilize predictions over large homogeneous regions. Although WaveMamba shows a noticeable improvement compared to Mamba, it remains inferior to HGFNet, implying that wavelet decomposition alone does not fully resolve the need for explicit global spatial--spectral coupling. This conclusion is visually supported by Fig. \ref{IPGT}, where HGFNet produces more compact and consistent class regions with fewer scattered errors, reflecting stronger structural regularization in the predicted map.

\begin{table}[!hbt]
    \centering
    \caption{Class-wise accuracy and overall performance metrics on \textbf{HH Dataset}.} 
    \resizebox{\columnwidth}{!}{\begin{tabular}{c||ccccc||c} \hline 
        \textbf{Class} & SF & HViT & WFormer & Mamba & WMamba & \textbf{HGFNet}  \\ \hline 
        1 & 99.4159 & 99.5584 & 99.7008 & 86.8660 & 95.1709 & 99.2592 \\ 
        2 & 93.6788 & 91.3439 & 94.9886 & 64.1799 & 88.3826 & 98.9749 \\ 
        3 & 97.7087 & 98.5519 & 99.6333 & 88.9011 & 93.3644 & 99.2484 \\ 
        4 & 99.7121 & 98.8033 & 99.8860 & 98.8731 & 99.7427 & 99.9693 \\ 
        5 & 91.9588 & 95.5934 & 96.1724 & 79.4467 & 76.5519 & 95.2717 \\ 
        6 & 98.2808 & 99.5466 & 99.8967 & 91.1845 & 98.9272 & 99.8788 \\ 
        7 & 97.1788 & 97.4278 & 98.3820 & 78.3189 & 87.7862 & 98.2326 \\ 
        8 & 89.7385 & 84.2624 & 86.9264 & 31.7217 & 63.0981 & 90.6265 \\ 
        9 & 99.4823 & 98.9646 & 97.1898 & 96.3948 & 94.3612 & 99.9075 \\ 
        10 & 97.4342 & 98.8058 & 98.6767 & 52.2672 & 87.2357 & 99.7256 \\ 
        11 & 98.2204 & 98.0025 & 98.2749 & 68.2585 & 83.5118 & 98.0388 \\ 
        12 & 89.0998 & 86.6205 & 93.4331 & 60.0402 & 78.2443 & 98.3247 \\ 
        13 & 97.8674 & 95.9747 & 92.9536 & 77.0481 & 83.2237 & 99.1025 \\ 
        14 & 97.7705 & 92.5231 & 98.5589 & 86.6503 & 97.1179 & 99.6193 \\ 
        15 & 94.8103 & 96.6067 & 99.2015 & 53.0938 & 39.9201 & 98.8023 \\ 
        16 & 99.5868 & 99.1462 & 96.0341 & 76.5078 & 95.0151 & 99.5042 \\ 
        17 & 96.2126 & 84.1196 & 99.1362 & 82.9900 & 80.1328 & 96.4119 \\ 
        18 & 98.3830 & 96.5174 & 94.8383 & 84.0796 & 85.0124 & 95.8333 \\ 
        19 & 97.8879 & 97.4288 & 98.1175 & 76.1019 & 90.1974 & 98.6914 \\ 
        20 & 96.0986 & 85.0258 & 92.7710 & 34.5381 & 88.9271 & 99.5410 \\ 
        21 & 94.1265 & 57.0783 & 70.7831 & 5.2710 & 54.3674 & 99.2469 \\ 
        22 & 99.2079 & 97.3762 & 99.7029 & 45.9405 & 93.8118 & 98.1683 \\ \hline \hline 

        \textbf{$\kappa$} & 97.8750 & 96.8637 & 98.0616 & 83.2685 & 92.3374 & 99.1331 \\ 
        \textbf{OA} & 98.3180 & 97.5143 & 98.4664 & 86.7859 & 93.9450 & 99.3141 \\ 
        \textbf{AA} & 96.5391 & 93.1490 & 95.6935 & 69.0306 & 84.2774 & 98.2900 \\ \hline 
    \end{tabular}}
    \label{Tab3}
\end{table}
\begin{figure}[!htb]
    \centering
    \begin{minipage}{0.060\textwidth}
        \centering
        \includegraphics[width=\textwidth]{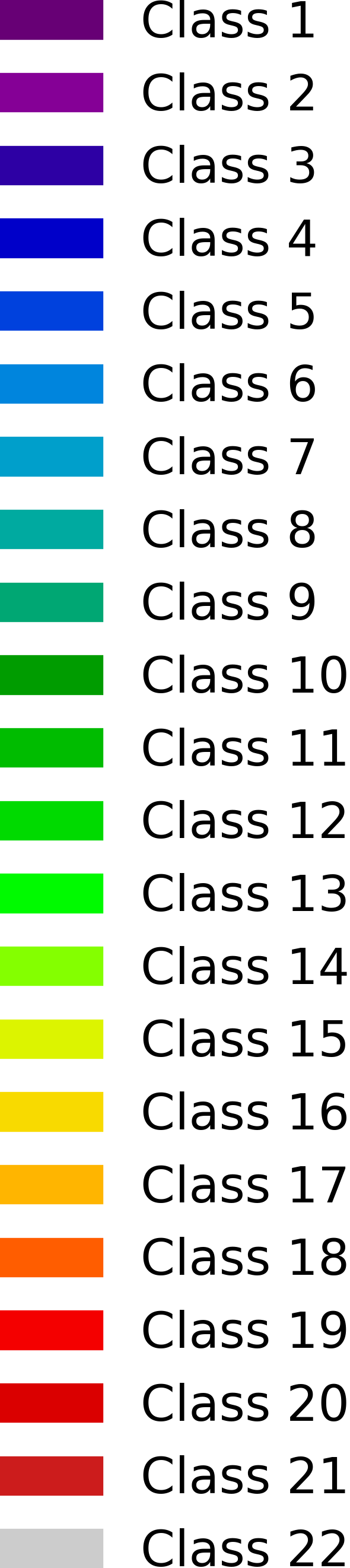}
    \end{minipage}
    \begin{minipage}{0.42\textwidth}
        \centering
        \begin{subfigure}{0.20\textwidth}
            \centering
            \includegraphics[width=0.99\textwidth]{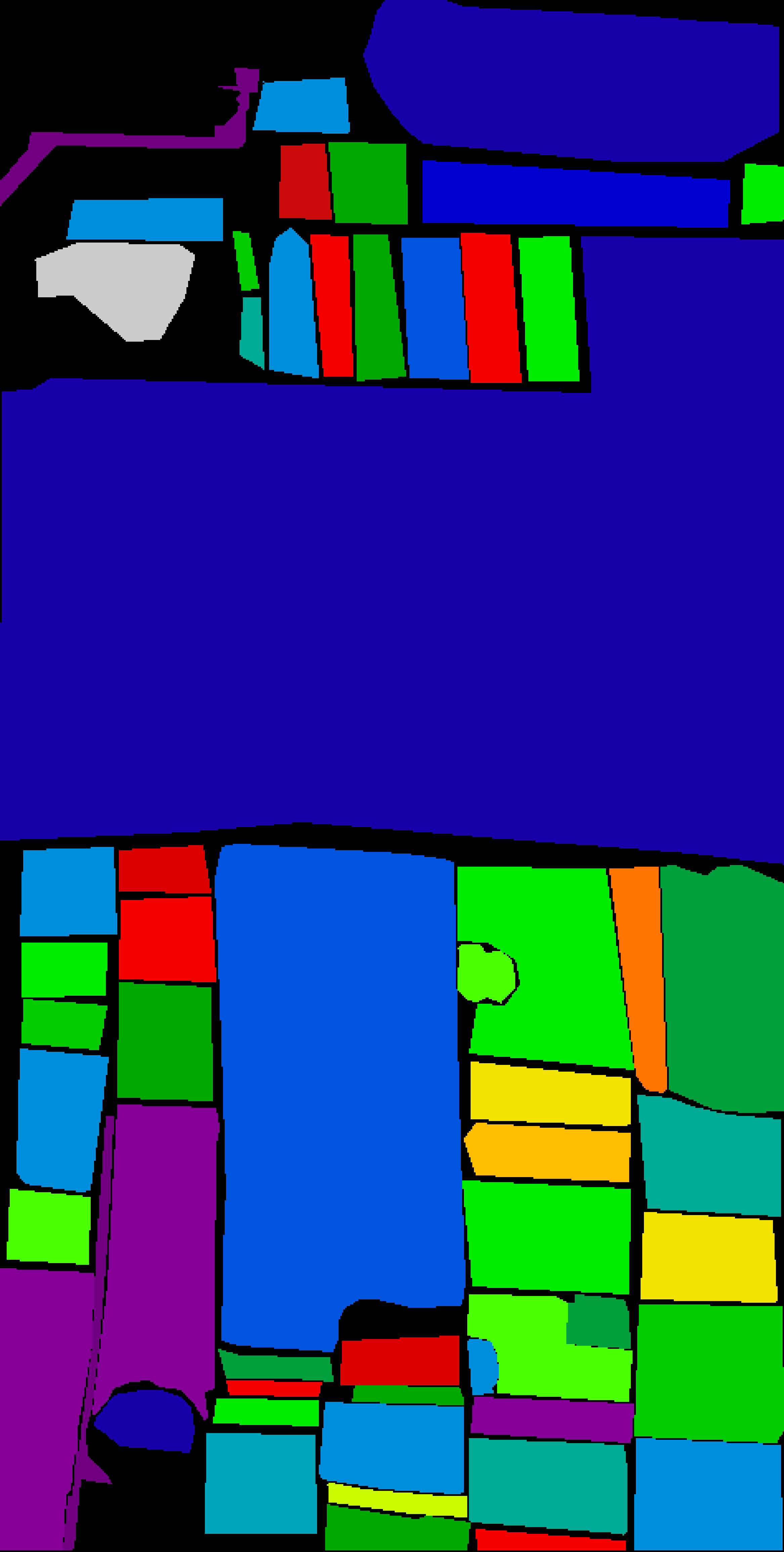}
            \caption*{GT}
        \end{subfigure}
        \begin{subfigure}{0.20\textwidth}
            \centering
            \includegraphics[width=0.99\textwidth]{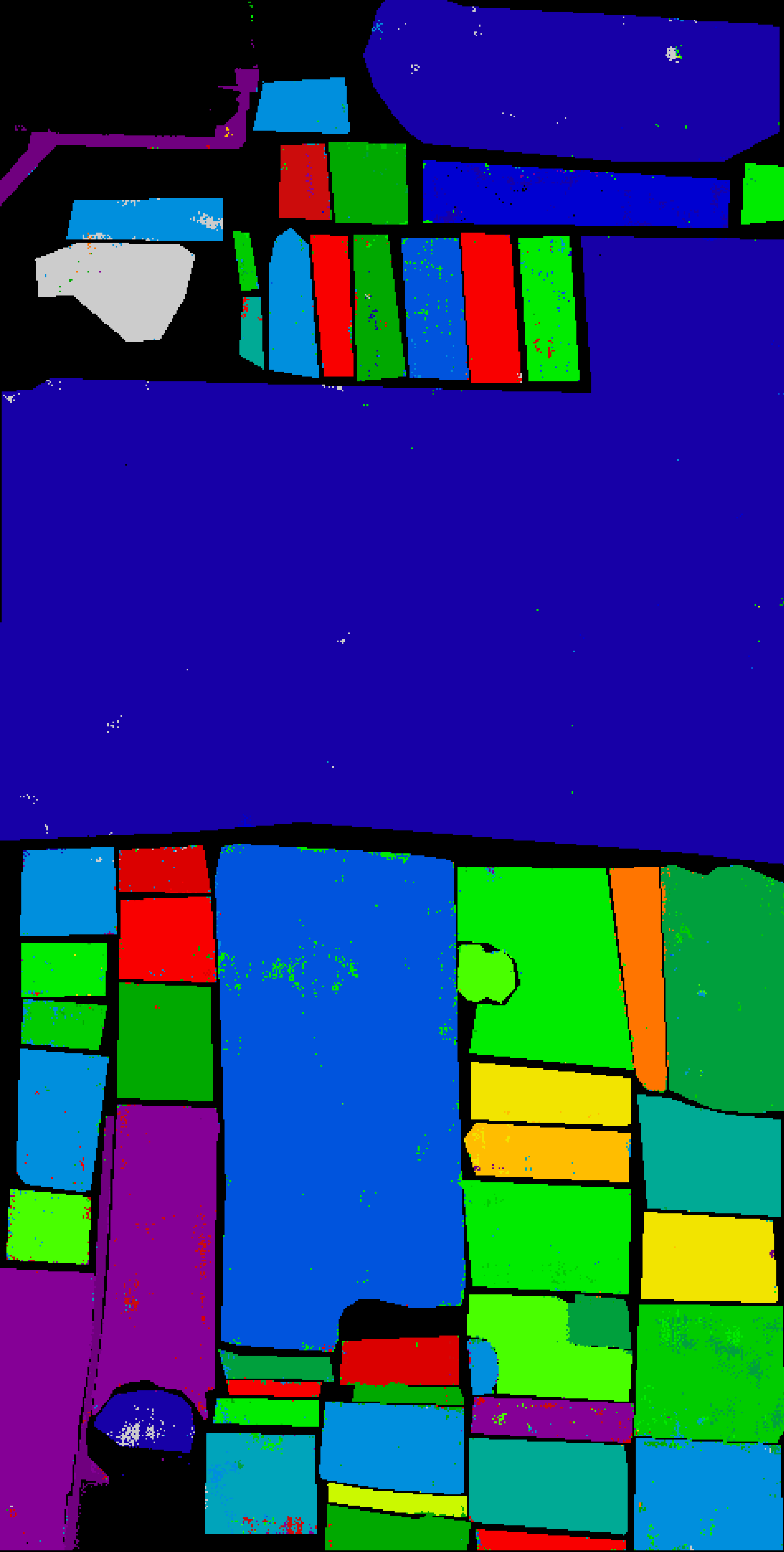}
            \caption*{SF}
        \end{subfigure}
        \begin{subfigure}{0.20\textwidth}
            \centering
            \includegraphics[width=0.99\textwidth]{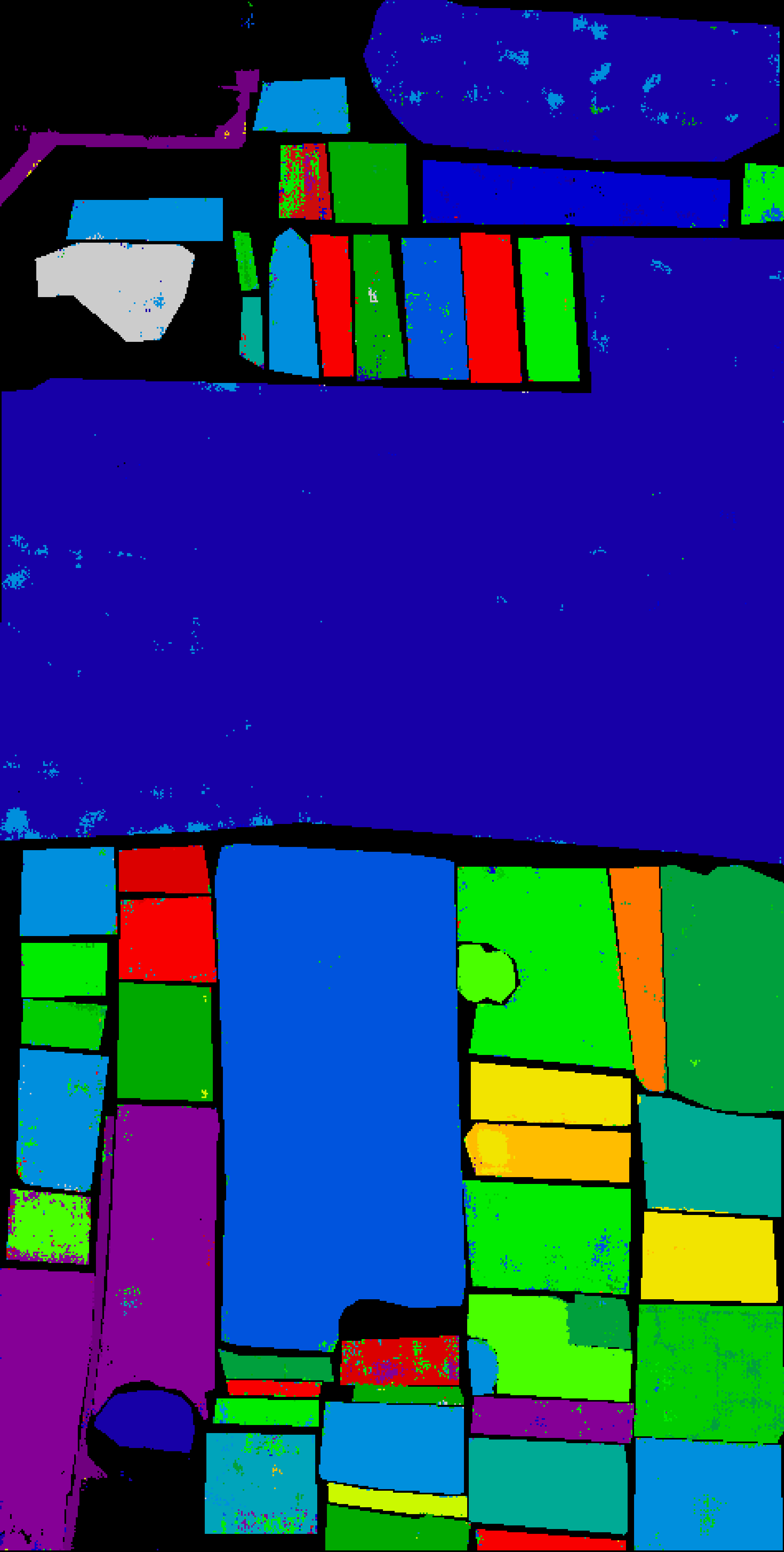}
            \caption*{HViT}
        \end{subfigure}
        \begin{subfigure}{0.20\textwidth}
            \includegraphics[width=0.99\textwidth]{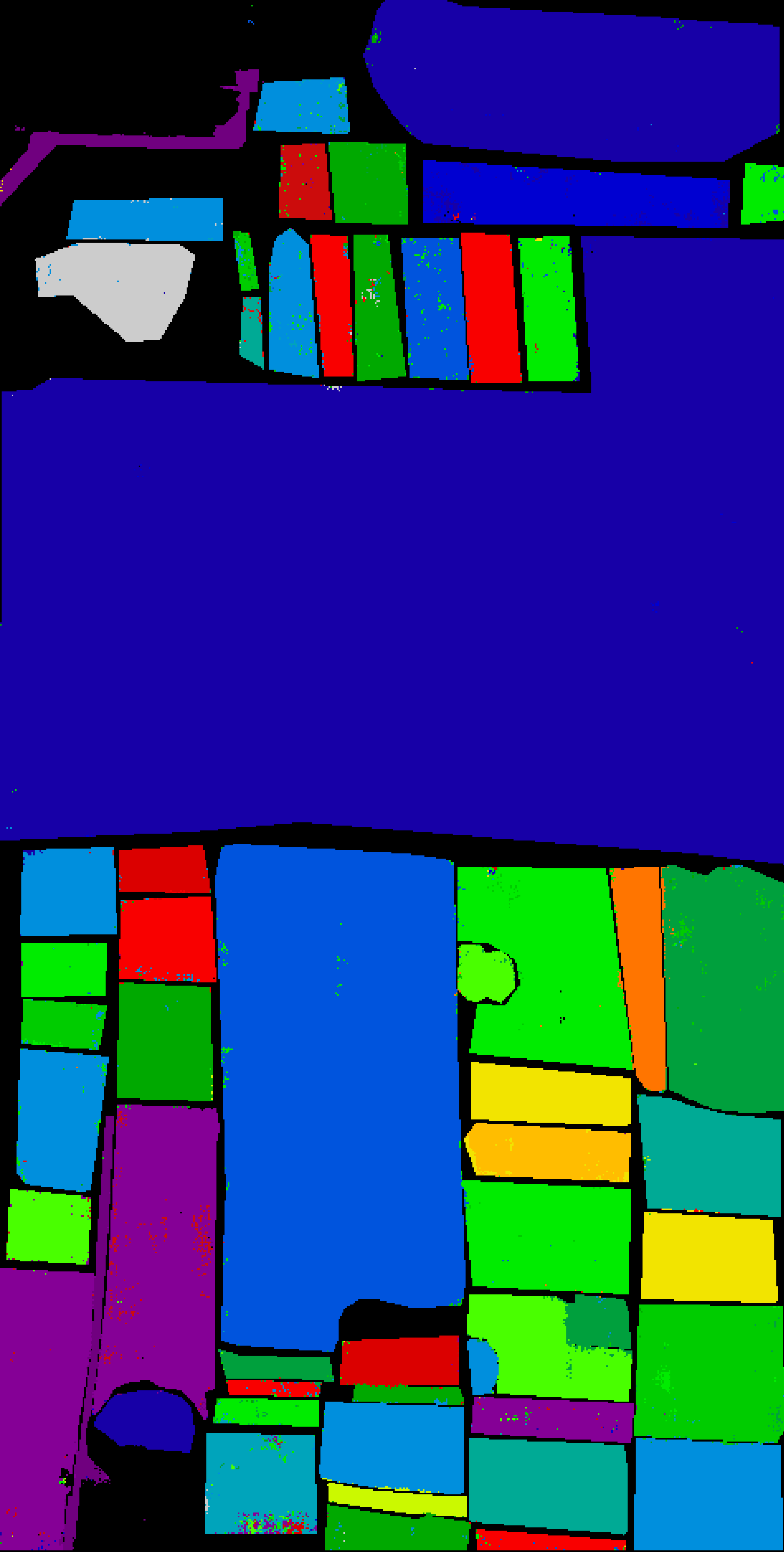}
            \caption*{WFormer}
        \end{subfigure}
        \begin{subfigure}{0.20\textwidth}
            \centering
            \includegraphics[width=0.99\textwidth]{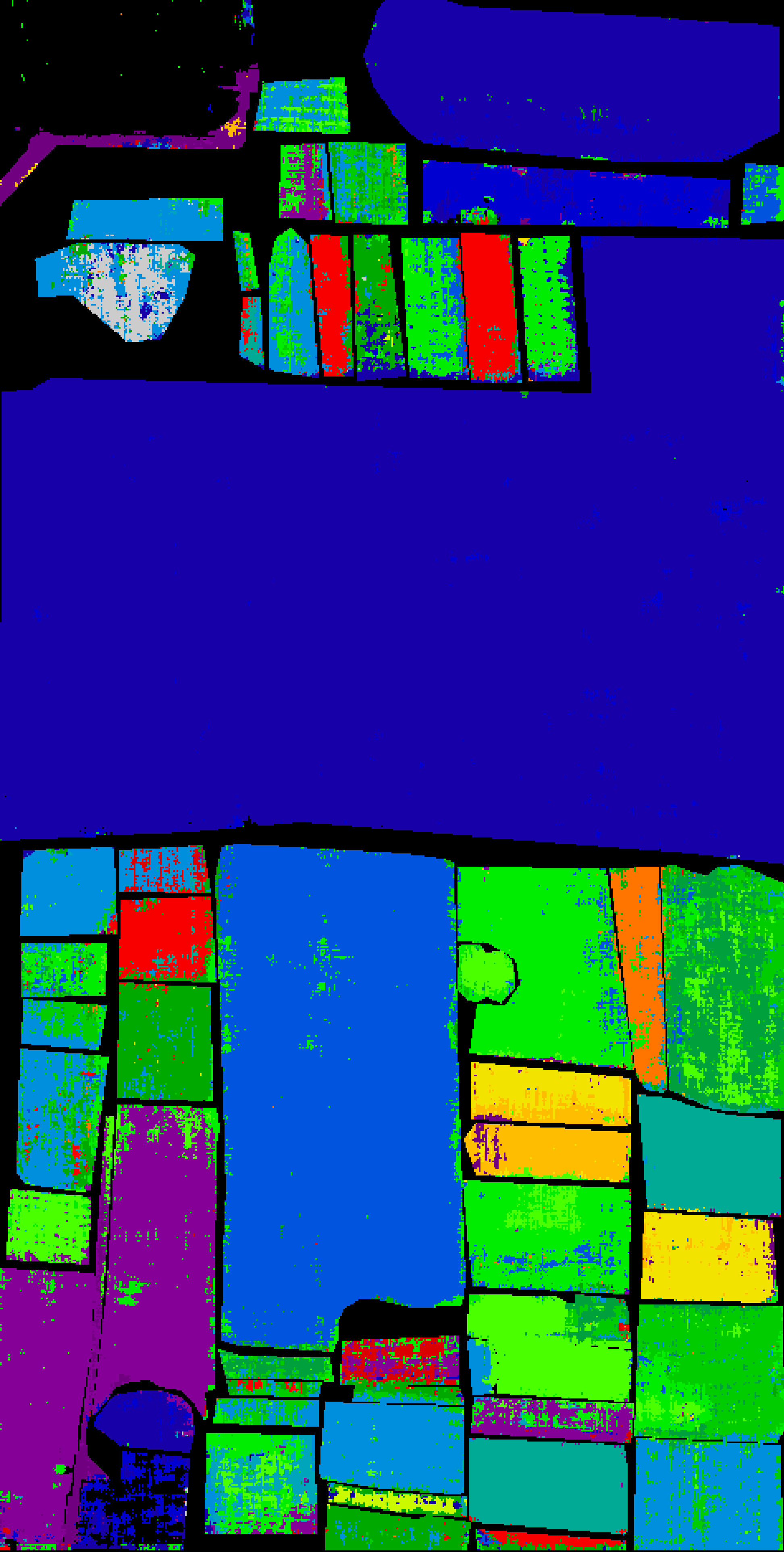}
            \caption*{Mamba}
        \end{subfigure}
        \begin{subfigure}{0.20\textwidth}
            \centering
            \includegraphics[width=0.99\textwidth]{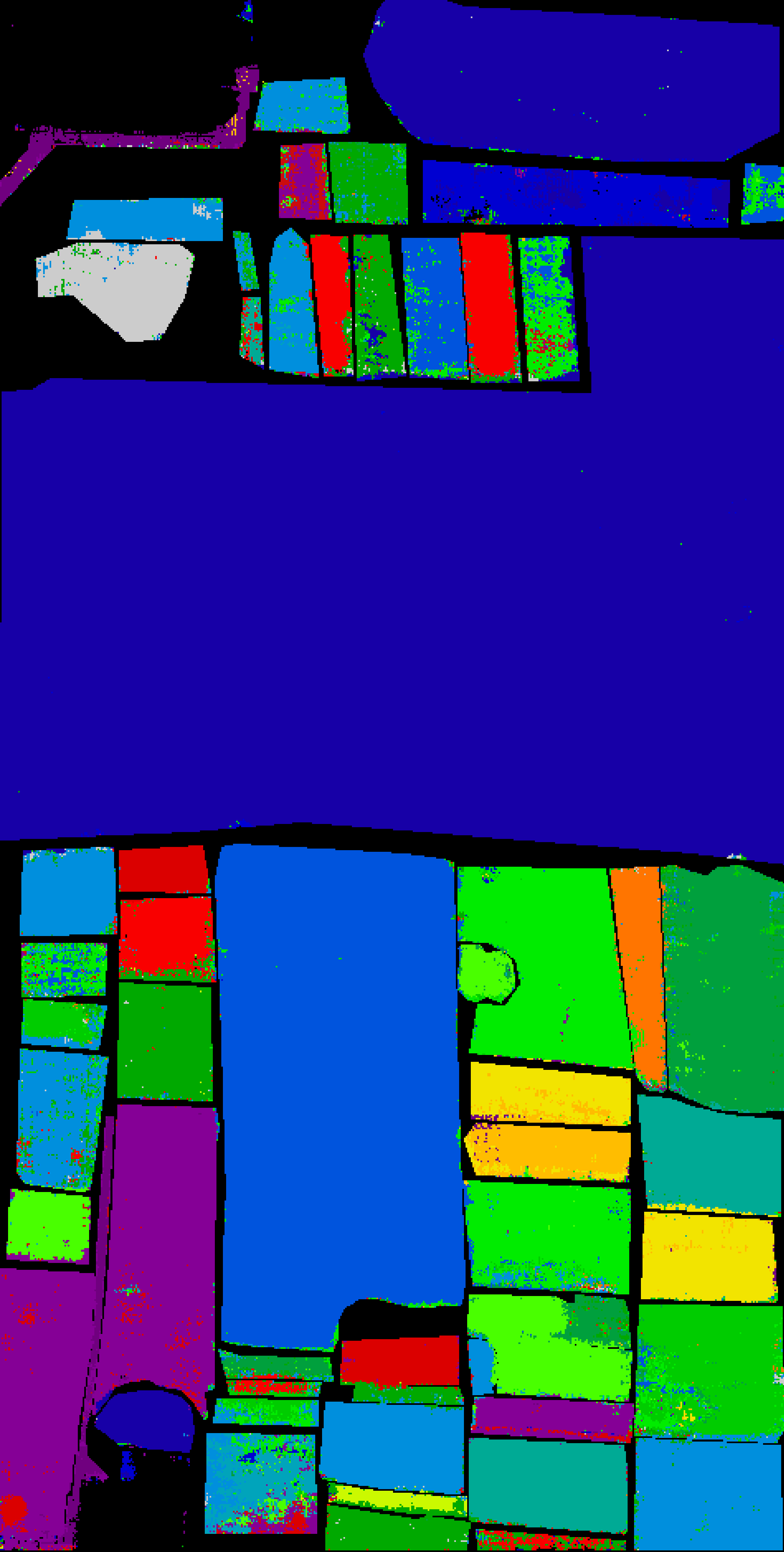}
            \caption*{WMamba}
        \end{subfigure}
        \begin{subfigure}{0.20\textwidth}
            \centering
            \includegraphics[width=0.99\textwidth]{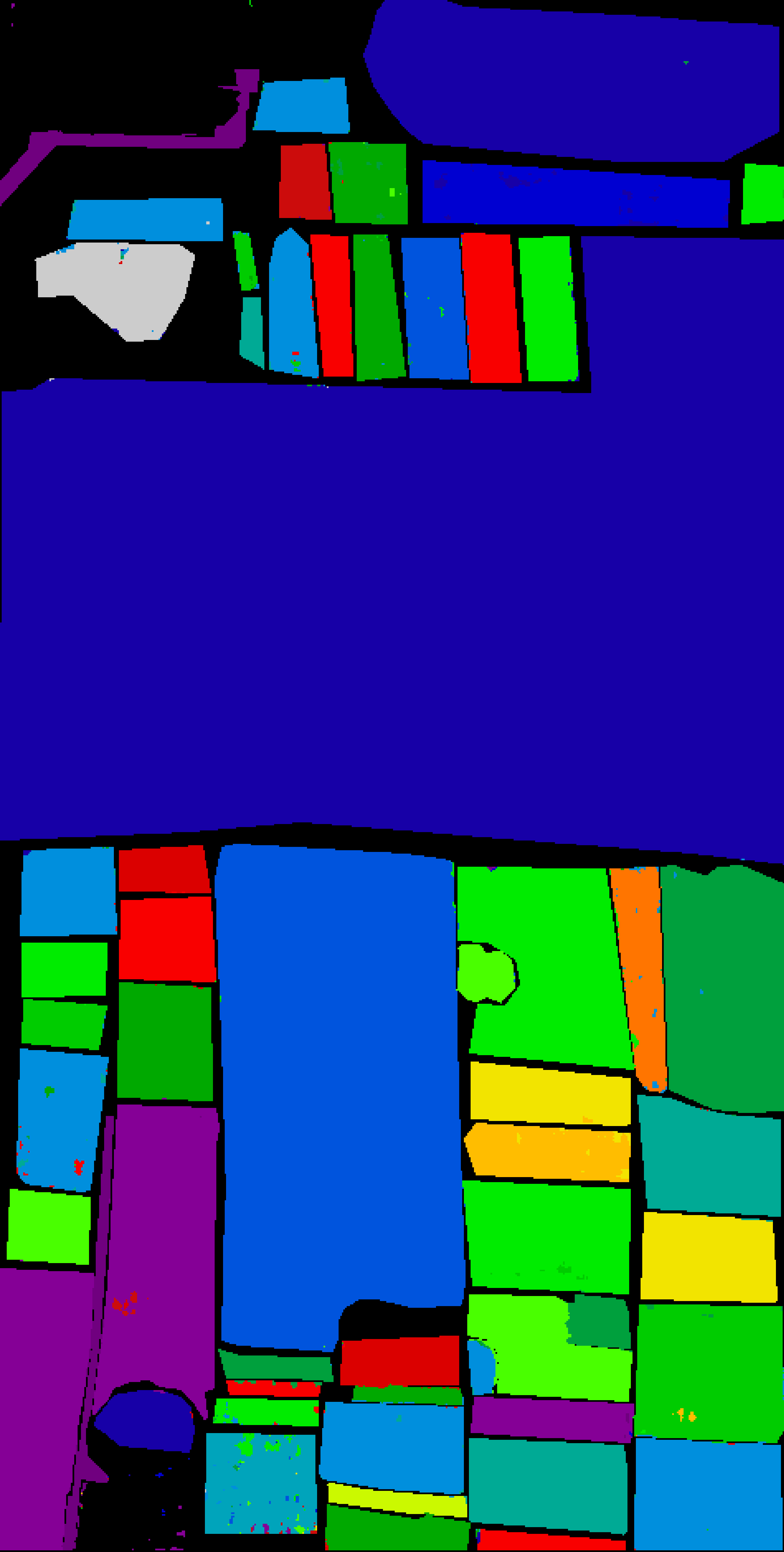}
            \caption*{HGFNet}
        \end{subfigure}
        \end{minipage}
    \caption{Predicted ground truth maps for the HH dataset.}
    \label{HHGT}
\end{figure}

Table \ref{Tab3} reports results on the HH dataset, where HGFNet achieves the highest OA ($99.3141\%$), $\kappa$ ($99.1331\%$), and AA ($98.2900\%$), confirming that the proposed architecture generalizes well across datasets with different spatial resolutions and class compositions. Compared with WaveFormer, which is the next best model in terms of OA, HGFNet yields a higher AA and more stable class-wise performance, indicating improved robustness across both frequent and minority classes. Notably, Mamba exhibits severe degradation in several classes (e.g., classes 8, 10, 15, and 21), which substantially lowers AA and reveals instability when discriminative cues require both local neighborhood evidence and globally consistent context. WaveMamba partially mitigates this issue, but still lags behind transformer-based methods and HGFNet, suggesting that combining a sequence model with wavelet priors is beneficial but not sufficient to match the stronger inductive bias produced by explicitly integrating 3D spatial-spectral locality with global Fourier mixing. The qualitative comparison in Fig. \ref{HHGT} is consistent with the metrics: HGFNet generates the cleanest and most spatially consistent map, with reduced boundary fragmentation and fewer misclassification streaks, indicating superior preservation of large-scale structures and more reliable class separation.

Overall, the results across HC, IP, and HH consistently demonstrate that HGFNet achieves state-of-the-art performance by unifying complementary mechanisms: the 3D CNN backbone captures local spatial-spectral interactions and suppresses pixel-level noise, while Fourier-based global feature extraction enables efficient long-range dependency modeling and improves global consistency. This combination is particularly advantageous in HSIC, where (i) many categories exhibit high spectral correlation, (ii) spatial context is critical for disambiguating mixed pixels and boundary regions, and (iii) limited labeled samples demand strong inductive priors to avoid overfitting. Consequently, HGFNet not only improves OA/$\kappa$/AA but also produces visually coherent classification maps with fewer artifacts, supporting its suitability for practical hyperspectral scene understanding.

\section{Conclusion and Future Research Directions}

This work proposed Hybrid GFNet (HGFNet), a novel hybrid architecture for hyperspectral image classification that explicitly unifies local spatial-spectral modeling with efficient global frequency-domain learning. By integrating a 3D convolutional front-end with GFNet-inspired global Fourier filtering blocks, HGFNet effectively captures fine-grained local structures as well as long-range spatial-spectral dependencies. The introduction of three complementary frequency operators, Spectral Fourier Transform (SFT), Spatial Fourier Transform (SpFT), and Spatial-Spectral Fourier Transform (SSFT), enables high-dimensional frequency modeling tailored to hyperspectral data, allowing the network to exploit inter-band spectral correlations and global spatial patterns simultaneously. In addition, the proposed Adaptive Focal Loss (AFL) explicitly addresses class imbalance by dynamically adjusting class-wise focusing behavior, while the adaptive fully connected classification head improves flexibility and generalization across datasets with diverse class distributions. Extensive experiments on multiple benchmark hyperspectral datasets demonstrate that HGFNet consistently outperforms state-of-the-art methods in terms of overall accuracy, average accuracy, and Cohen’s kappa, while also producing more spatially coherent and visually consistent classification maps.

Despite its strong empirical performance, several limitations remain and motivate future research. First, the reliance on Fourier-based global feature extraction inherently assumes signal stationarity, which may limit the model’s ability to adapt to highly non-stationary spectral variations caused by complex illumination conditions, atmospheric effects, and sensor noise in real-world hyperspectral imagery. Incorporating adaptive and localized frequency representations, such as learnable spectral bases and hybrid Fourier–wavelet formulations, could improve robustness under such conditions. Second, although the 3D convolutional layers provide strong local inductive biases, their computational and memory costs may hinder scalability to ultra-high-resolution hyperspectral scenes or large-scale datasets. Future work could explore lightweight alternatives, such as depthwise separable 3D convolutions or dynamic kernel decomposition, to further reduce complexity without losing representational power. Finally, while AFL improves performance under moderate class imbalance, extremely skewed class distributions may require more advanced strategies, such as curriculum-based reweighting, meta-learning of loss functions, or joint sampling–loss optimization. Addressing these challenges will further enhance the practicality and generalization capability of HGFNet, paving the way for robust deployment in large-scale and real-world hyperspectral applications.

\bibliographystyle{IEEEtran}
\bibliography{Sam}

@article{TULU2026111273,
title = {RGB-to-synthetic-thermal image translation using generative AI to support crop water stress assessment},
journal = {Computers and Electronics in Agriculture},
volume = {241},
pages = {111273},
year = {2026},
issn = {0168-1699},
doi = {https://doi.org/10.1016/j.compag.2025.111273},
url = {https://www.sciencedirect.com/science/article/pii/S0168169925013791},
author = {Boaz B. Tulu and Fitsum T. Teshome and Yiannis Ampatzidis and Changying Li and Willingthon Pavan and Golmar Golmohammadi and Haimanote K. Bayabil}}

@ARTICLE{11392780,
  author={Ahmad, Faiq and Usama, Muhammad and Ghous, Usman and Shehzad, Danish and Mazzara, Manuel and Ahmad, Muhammad},
  journal={IEEE Geoscience and Remote Sensing Letters}, 
  title={A Spiking and Memory-Enhanced State-Space Model for Hyperspectral Image Classification}, 
  year={2026},
  volume={23},
  number={},
  pages={1-5},
  doi={10.1109/LGRS.2026.3663905}}

@ARTICLE{11119702,
  author={Sohail, Saad and Usama, Muhammad and Ghous, Usman and Mazzara, Manuel and Distefano, Salvatore and Ahmad, Muhammad},
  journal={IEEE Geoscience and Remote Sensing Letters}, 
  title={EnergyFormer: Energy Attention With Fourier Embedding for Hyperspectral Image Classification}, 
  year={2025},
  volume={22},
  number={},
  pages={1-5},
  doi={10.1109/LGRS.2025.3596629}}

@ARTICLE{11105087,
  author={Ahmad, Muhammad and Mauro, Francesco and Raza, Rana Aamir and Mazzara, Manuel and Distefano, Salvatore and Khan, Adil Mehmood and Ullo, Silvia Liberata},
  journal={IEEE Journal of Selected Topics in Applied Earth Observations and Remote Sensing}, 
  title={Transformer-Driven Active Transfer Learning for Cross-Hyperspectral Image Classification}, 
  year={2025},
  volume={18},
  number={},
  pages={19635-19648},
  doi={10.1109/JSTARS.2025.3594108}}

@article{Ahmad18072025,
author = {Muhammad Ahmad and Manuel Mazzara and Salvatore Distefano and Adil Mehmood Khan and Xin Wu},
title = {Self-supervised spatial-spectral transformer with Extreme Learning Machine for Hyperspectral Image Classification},
journal = {International Journal of Remote Sensing},
volume = {46},
number = {14},
pages = {5384--5407},
year = {2025},
publisher = {Taylor \& Francis},
doi = {10.1080/01431161.2025.2520049},
URL = {https://doi.org/10.1080/01431161.2025.2520049},
eprint = {https://doi.org/10.1080/01431161.2025.2520049}}

@ARTICLE{11037730,
  author={Sohail, Saad and Usama, Muhammad and Ghous, Usman and Mazzara, Manuel and Ahmad, Muhammad},
  journal={IEEE Geoscience and Remote Sensing Letters}, 
  title={Differential Attention With Enhanced Squeeze-and-Excitation for Hyperspectral Image Classification}, 
  year={2025},
  volume={22},
  number={},
  pages={1-5},
  doi={10.1109/LGRS.2025.3580630}}

@ARTICLE{10955699,
  author={Ahmad, Muhammad and Mazzara, Manuel and Distefano, Salvatore and Khan, Adil Mehmood and Ullo, Silvia Liberata},
  journal={IEEE Journal of Selected Topics in Applied Earth Observations and Remote Sensing}, 
  title={DiffFormer: A Differential Spatial-Spectral Transformer for Hyperspectral Image Classification}, 
  year={2025},
  volume={18},
  number={},
  pages={10419-10428},
  doi={10.1109/JSTARS.2025.3558889}}

@ARTICLE{9767615,
  author={Ahmad, Muhammad and Khan, Adil Mehmood and Mazzara, Manuel and Distefano, Salvatore and Roy, Swalpa Kumar and Wu, Xin},
  journal={IEEE Journal of Selected Topics in Applied Earth Observations and Remote Sensing}, 
  title={Hybrid Dense Network With Attention Mechanism for Hyperspectral Image Classification}, 
  year={2022},
  volume={15},
  number={},
  pages={3948-3957},
  doi={10.1109/JSTARS.2022.3171586}}

@article{KARUKAYIL2026111282,
title = {3D crop reconstruction: A review of hyperspectral and multispectral approaches},
journal = {Computers and Electronics in Agriculture},
volume = {241},
pages = {111282},
year = {2026},
issn = {0168-1699},
doi = {https://doi.org/10.1016/j.compag.2025.111282},
url = {https://www.sciencedirect.com/science/article/pii/S0168169925013882},
author = {Abhiram Karukayil and João F.C. Mota and Fernando Auat Cheein}}

@article{SERRANTI2026129361,
title = {Comparison of Hyperspectral Imaging and FTIR Spectroscopy for Microplastic Polymer Identification: Proposal of a Scalable Protocol Validated in a 12-Month River Survey},
journal = {Talanta},
pages = {129361},
year = {2026},
issn = {0039-9140},
doi = {https://doi.org/10.1016/j.talanta.2026.129361},
url = {https://www.sciencedirect.com/science/article/pii/S0039914026000147},
author = {Silvia Serranti and Giuseppe Bonifazi and Pietro Cocozza and Paola Cucuzza and Roberta Palmieri and Margherita Benzi and Marco Lezzi and Cristina Mazziotti and Elena Riccardi and Elena Barbieri and Irene Ingrando and Fernanda Moroni}}

@article{LKHAOUA2026109997,
title = {Proximal mapping of carbonate-fluorapatite using hyperspectral imaging with SEM-EDS mineralogical validation: a drill core case study from the Ben Guerir Mine, Morocco},
journal = {Minerals Engineering},
volume = {237},
pages = {109997},
year = {2026},
issn = {0892-6875},
doi = {https://doi.org/10.1016/j.mineng.2025.109997},
url = {https://www.sciencedirect.com/science/article/pii/S0892687525008258},
author = {Houda Lkhaoua and Otmane Raji and Abdellatif Elghali and Abdelhafid EL {Alaoui EL Fels} and Mohamed Mazigh and Mostafa Benzaazoua}}

@ARTICLE{11222092,
  author={Ahmad, Muhammad and Mazzara, Manuel and Distefano, Salvatore and Khan, Adil Mehmood},
  journal={IEEE Transactions on Geoscience and Remote Sensing}, 
  title={Byte Latent Mamba With State Space and Knowledge Distillation for Hyperspectral Image Classification}, 
  year={2025},
  volume={63},
  number={},
  pages={1-15},
  doi={10.1109/TGRS.2025.3626861}}

@article{Wang26,
author = {Wang, Baosheng and Ping, Xiaoxue and Liu, Yang},
title = {Hyperspectral System Coupled With a Global Spectral Feature Classification Network to Identify the Origin of Mung Bean},
journal = {Journal of Food Science},
volume = {91},
number = {1},
pages = {e70813},
doi = {https://doi.org/10.1111/1750-3841.70813},
url = {https://ift.onlinelibrary.wiley.com/doi/abs/10.1111/1750-3841.70813},
eprint = {https://ift.onlinelibrary.wiley.com/doi/pdf/10.1111/1750-3841.70813},
year = {2026}
}

@ARTICLE{11226902,
  author={Ahmad, Muhammad and Mazzara, Manuel and Distefano, Salvatore and Khan, Adil Mehmood and Butt, Muhammad Hassaan Farooq and Usama, Muhammad and Hong, Danfeng},
  journal={IEEE Transactions on Emerging Topics in Computing}, 
  title={GraphMamba: Graph Tokenization Mamba for Hyperspectral Image Classification}, 
  year={2025},
  volume={13},
  number={4},
  pages={1510-1521},
  doi={10.1109/TETC.2025.3626943}}

@ARTICLE{11090003,
  author={Ahmad, Muhammad and Mazzara, Manuel and Distefano, Salvatore and Mehmood Khan, Adil and Hassaan Farooq Butt, Muhammad and Hong, Danfeng},
  journal={IEEE Transactions on Neural Networks and Learning Systems}, 
  title={PolicyMamba: Localized Policy Attention With State Space Model for Land Cover Classification}, 
  year={2025},
  volume={36},
  number={10},
  pages={17814-17825},
  doi={10.1109/TNNLS.2025.3586836}}

@article{Ahmad03042025,
author = {Muhammad Ahmad and Muhammad Hassaan Farooq Butt and Muhammad Usama and Hamad Ahmed Altuwaijri and Manuel Mazzara and Salvatore Distefano and Adil Mehmood Khan and},
title = {Multi-head spatial-spectral mamba for hyperspectral image classification},
journal = {Remote Sensing Letters},
volume = {16},
number = {4},
pages = {339--353},
year = {2025},
publisher = {Taylor \& Francis},
doi = {10.1080/2150704X.2025.2461330},
URL = {https://doi.org/10.1080/2150704X.2025.2461330},
eprint = {https://doi.org/10.1080/2150704X.2025.2461330}}

@ARTICLE{10483019,
  author={Lin, Jiacheng and Chen, Jiajun and Yang, Kailun and Roitberg, Alina and Li, Siyu and Li, Zhiyong and Li, Shutao},
  journal={IEEE Transactions on Neural Networks and Learning Systems}, 
  title={AdaptiveClick: Click-Aware Transformer With Adaptive Focal Loss for Interactive Image Segmentation}, 
  year={2025},
  volume={36},
  number={3},
  pages={5759-5773},
  doi={10.1109/TNNLS.2024.3378295}}

@ARTICLE{9627165,
  author={Hong, Danfeng and Han, Zhu and Yao, Jing and Gao, Lianru and Zhang, Bing and Plaza, Antonio and Chanussot, Jocelyn},
  journal={IEEE Transactions on Geoscience and Remote Sensing}, 
  title={SpectralFormer: Rethinking Hyperspectral Image Classification With Transformers}, 
  year={2022},
  volume={60},
  number={},
  pages={1-15},
  doi={10.1109/TGRS.2021.3130716}}

@ARTICLE{10767233,
  author={Ahmad, Muhammad and Usama, Muhammad and Mazzara, Manuel and Distefano, Salvatore},
  journal={IEEE Geoscience and Remote Sensing Letters}, 
  title={WaveMamba: Spatial-Spectral Wavelet Mamba for Hyperspectral Image Classification}, 
  year={2025},
  volume={22},
  number={},
  pages={1-5},
  doi={10.1109/LGRS.2024.3506034}}

@ARTICLE{10604894,
  author={Li, Yapeng and Luo, Yong and Zhang, Lefei and Wang, Zengmao and Du, Bo},
  journal={IEEE Transactions on Geoscience and Remote Sensing}, 
  title={MambaHSI: Spatial–Spectral Mamba for Hyperspectral Image Classification}, 
  year={2024},
  volume={62},
  number={},
  pages={1-16},
  doi={10.1109/TGRS.2024.3430985}}

@ARTICLE{10399798,
  author={Ahmad, Muhammad and Ghous, Usman and Usama, Muhammad and Mazzara, Manuel},
  journal={IEEE Geoscience and Remote Sensing Letters}, 
  title={WaveFormer: Spectral–Spatial Wavelet Transformer for Hyperspectral Image Classification}, 
  year={2024},
  volume={21},
  number={},
  pages={1-5},
  doi={10.1109/LGRS.2024.3353909}}

@article{khan2024deep,
  title={Deep spectral spatial feature enhancement through transformer for hyperspectral image classification},
  author={Khan, Rahim and Arshad, Tahir and Ma, Xuefei and Chen, Wang and Haifeng, Zhu and Yanni, Wu},
  journal={IEEE Geoscience And Remote Sensing Letters},
  year={2024},
  publisher={IEEE}
}

@article{AHMAD2025130428,
title = {A comprehensive survey for Hyperspectral Image Classification: The evolution from conventional to transformers and Mamba models},
journal = {Neurocomputing},
volume = {644},
pages = {130428},
year = {2025},
issn = {0925-2312},
doi = {https://doi.org/10.1016/j.neucom.2025.130428},
url = {https://www.sciencedirect.com/science/article/pii/S0925231225011002},
author = {Muhammad Ahmad and Salvatore Distefano and Adil Mehmood Khan and Manuel Mazzara and Chenyu Li and Hao Li and Jagannath Aryal and Yao Ding and Gemine Vivone and Danfeng Hong}}

@ARTICLE{9903062,
  author={Ahmad, Muhammad and Ghous, Usman and Hong, Danfeng and Khan, Adil Mehmood and Yao, Jing and Wang, Shaohua and Chanussot, Jocelyn},
  journal={IEEE Transactions on Geoscience and Remote Sensing}, 
  title={A Disjoint Samples-Based 3D-CNN With Active Transfer Learning for Hyperspectral Image Classification}, 
  year={2022},
  volume={60},
  number={},
  pages={1-16},
  doi={10.1109/TGRS.2022.3209182}}

@ARTICLE{10604879,
  author={Ahmad, Muhammad and Usama, Muhammad and Khan, Adil Mehmood and Distefano, Salvatore and Altuwaijri, Hamad Ahmed and Mazzara, Manuel},
  journal={IEEE Geoscience and Remote Sensing Letters}, 
  title={Spatial–Spectral Transformer With Conditional Position Encoding for Hyperspectral Image Classification}, 
  year={2024},
  volume={21},
  number={},
  pages={1-5},
  doi={10.1109/LGRS.2024.3431188}}

@ARTICLE{10685113,
  author={Ahmad, Muhammad and Usama, Muhammad and Mazzara, Manuel and Distefano, Salvatore and Altuwaijri, Hamad Ahmed and Ullo, Silvia Liberata},
  journal={IEEE Journal of Selected Topics in Applied Earth Observations and Remote Sensing}, 
  title={Fusing Transformers in a Tuning Fork Structure for Hyperspectral Image Classification Across Disjoint Samples}, 
  year={2024},
  volume={17},
  number={},
  pages={18167-18181},
  doi={10.1109/JSTARS.2024.3465831}}

@ARTICLE{10838328,
  author={Ming, Ri and Chen, Na and Peng, Jiangtao and Sun, Weiwei and Ye, Zhijing},
  journal={IEEE Journal of Selected Topics in Applied Earth Observations and Remote Sensing}, 
  title={Semantic Tokenization-Based Mamba for Hyperspectral Image Classification}, 
  year={2025},
  volume={18},
  number={},
  pages={4227-4241},
  doi={10.1109/JSTARS.2025.3528122}}

@ARTICLE{10472541,
  author={Zhao, Zhuoyi and Xu, Xiang and Li, Shutao and Plaza, Antonio},
  journal={IEEE Transactions on Geoscience and Remote Sensing}, 
  title={Hyperspectral Image Classification Using Groupwise Separable Convolutional Vision Transformer Network}, 
  year={2024},
  volume={62},
  number={},
  pages={1-17},
  doi={10.1109/TGRS.2024.3377610}}

@ARTICLE{10731855,
  author={Chen, Liang and He, Jingfei and Shi, Hao and Yang, Jingyi and Li, Wei},
  journal={IEEE Transactions on Geoscience and Remote Sensing}, 
  title={SWDiff: Stage-Wise Hyperspectral Diffusion Model for Hyperspectral Image Classification}, 
  year={2024},
  volume={62},
  number={},
  pages={1-17},
  doi={10.1109/TGRS.2024.3485483}}

@ARTICLE{10628006,
  author={Yang, Teng and Xiao, Song and Qu, Jiahui and Dong, Wenqian and Du, Qian and Li, Yunsong},
  journal={IEEE Transactions on Image Processing}, 
  title={Graph Embedding Interclass Relation-Aware Adaptive Network for Cross-Scene Classification of Multisource Remote Sensing Data}, 
  year={2024},
  volume={33},
  number={},
  pages={4459-4474},
  doi={10.1109/TIP.2024.3422881}}

@ARTICLE{10746459,
  author={Yang, Aitao and Li, Min and Ding, Yao and Fang, Leyuan and Cai, Yaoming and He, Yujie},
  journal={IEEE Transactions on Geoscience and Remote Sensing}, 
  title={GraphMamba: An Efficient Graph Structure Learning Vision Mamba for Hyperspectral Image Classification}, 
  year={2024},
  volume={62},
  number={},
  pages={1-14},
  doi={10.1109/TGRS.2024.3493101}}

@article{SARPONG2024122202,
title = {Hyperspectral image classification using Second-Order Pooling with Graph Residual Unit Network},
journal = {Expert Systems with Applications},
volume = {238},
pages = {122202},
year = {2024},
issn = {0957-4174},
doi = {https://doi.org/10.1016/j.eswa.2023.122202},
url = {https://www.sciencedirect.com/science/article/pii/S0957417423027045},
author = {Kwabena Sarpong and Zhiguang Qin and Rajab Ssemwogerere and Rutherford Agbeshi Patamia and Asha Mzee Khamis and Enoch Opanin Gyamfi and Favour Ekong and Chiagoziem C. Ukwuoma}}

@ARTICLE{10509762,
  author={Chen, Shih-Yu and Hsu, Kai-Hsun and Kuo, Tzu-Hsien},
  journal={IEEE Journal of Selected Topics in Applied Earth Observations and Remote Sensing}, 
  title={Hyperspectral Target Detection-Based 2-D–3-D Parallel Convolutional Neural Networks for Hyperspectral Image Classification}, 
  year={2024},
  volume={17},
  number={},
  pages={9451-9469},
  doi={10.1109/JSTARS.2024.3394704}}

@ARTICLE{10493162,
  author={Shi, Hao and Zhang, Youqiang and Cao, Guo and Yang, Di},
  journal={IEEE Transactions on Geoscience and Remote Sensing}, 
  title={Fortifying Centers and Edges: Multidomain Feature Learning Meets Hyperspectral Image Classification}, 
  year={2024},
  volume={62},
  number={},
  pages={1-16},
  doi={10.1109/TGRS.2024.3385501}}

@ARTICLE{10847802,
  author={Li, Xin and Xu, Feng and Yu, Anzhu and Lyu, Xin and Gao, Hongmin and Zhou, Jun},
  journal={IEEE Transactions on Geoscience and Remote Sensing}, 
  title={A Frequency Decoupling Network for Semantic Segmentation of Remote Sensing Images}, 
  year={2025},
  volume={63},
  number={},
  pages={1-21},
  doi={10.1109/TGRS.2025.3531879}}

@ARTICLE{10091201,
  author={Rao, Yongming and Zhao, Wenliang and Zhu, Zheng and Zhou, Jie and Lu, Jiwen},
  journal={IEEE Transactions on Pattern Analysis and Machine Intelligence}, 
  title={GFNet: Global Filter Networks for Visual Recognition}, 
  year={2023},
  volume={45},
  number={9},
  pages={10960-10973},
  doi={10.1109/TPAMI.2023.3263824}}

@ARTICLE{10850760,
  author={Wu, Qinggang and He, Mengkun and Chen, Qiqiang and Sun, Le and Ma, Chao},
  journal={IEEE Journal of Selected Topics in Applied Earth Observations and Remote Sensing}, 
  title={Integrating Multi-Scale Spatial-Spectral Shuffling Convolution with 3D Lightweight Transformer for Hyperspectral Image Classification}, 
  year={2025},
  volume={},
  number={},
  pages={1-17},
  doi={10.1109/JSTARS.2025.3533211}}

@ARTICLE{10855493,
  author={Wang, Yiqun and Yang, Lina and Wu, Thomas Xinzhang and Tang, Kaiwen and Zha, Wanxing},
  journal={IEEE Transactions on Geoscience and Remote Sensing}, 
  title={Spatial-Spectral Fusion BiFormer: A Novel Dynamic Routing Approach for Hyperspectral Image Classification}, 
  year={2025},
  volume={},
  number={},
  pages={1-1},
  doi={10.1109/TGRS.2025.3534790}}

@ARTICLE{10820058,
  author={Fang, Yu and Sun, Le and Zheng, Yuhui and Wu, Zebin},
  journal={IEEE Transactions on Image Processing}, 
  title={Deformable Convolution-Enhanced Hierarchical Transformer With Spectral-Spatial Cluster Attention for Hyperspectral Image Classification}, 
  year={2025},
  volume={34},
  number={},
  pages={701-716},
  doi={10.1109/TIP.2024.3522809}}

@ARTICLE{10843260,
  author={Xi, Bobo and Zhang, Yun and Li, Jiaojiao and Zheng, Tie and Zhao, Xunfeng and Xu, Haitao and Xue, Changbin and Li, Yunsong and Chanussot, Jocelyn},
  journal={IEEE Transactions on Geoscience and Remote Sensing}, 
  title={MCTGCL: Mixed CNN–Transformer for Mars Hyperspectral Image Classification With Graph Contrastive Learning}, 
  year={2025},
  volume={63},
  number={},
  pages={1-14},
  doi={10.1109/TGRS.2025.3529996}}

\end{document}